\documentclass[a4,notitlepage,12pt]{jedm}
\usepackage[table]{xcolor}
\usepackage{url}
\usepackage{graphicx}
\usepackage{booktabs}

\usepackage{amsmath, amssymb}
\usepackage[normalem]{ulem} 
\usepackage{rotating} 

\begin{document}

\title{Predicting Performance During Tutoring with Models of Recent Performance}
\date{}

\author{{\large April Galyardt}\\ University of Georgia\\galyardt@uga.edu \and
 {\large Ilya Goldin}\\ Pearson \\ilya.goldin@pearson.com }

\maketitle

\begin{abstract}
In educational technology and learning sciences, there are multiple uses for a predictive model of whether a student will perform a task correctly or not. For example, an intelligent tutoring system may use such a model to estimate whether or not a student has mastered a skill. We analyze the significance of data recency in making such predictions, i.e., asking whether relatively more recent observations of a student's performance matter more than relatively older observations. We develop a new Recent-Performance Factors Analysis model that takes data recency into account. The new model significantly improves predictive accuracy over both existing logistic-regression performance models and over novel baseline models in evaluations on real-world and synthetic datasets. As a secondary contribution, we demonstrate how the widely used cross-validation with 0-1 loss is inferior to AIC and to cross-validation with $L_1$ prediction error loss as a measure of model performance.

\end{abstract}

\section{Introduction}
A central field of research in educational technology and assessment is concerned with modeling the probability that a student will respond correctly to some question. This modeling is used to analyze test answers, as with Item Response Theory; in adaptive learning technologies, such as the use of Bayesian Knowledge Tracing \cite{corbett_knowledge_1995} in intelligent tutoring systems; to analyze the domains that students study, such as the study of transfer across tasks \cite{Pavlik11}; and to understand student behaviors like gaming the system \cite{baker_off-task_2004}. 

Our work advances this field by examining alternative representations of recency. The intuition is simple: as students practice a skill, we expect their understanding to increase and their performance to improve. Having recently succeeded at a task may make it more likely that learning has taken place, and such a moment of learning ought to contribute to our prediction of successful performance. This work is the first thorough investigation of recency effects in performance modeling.

We begin by describing a space of models of recency that fits into the logistic regression approach to performance modeling, as exemplified by Item Response Theory models. This space subsumes many existing modeling efforts, including the Additive Factors Model (AFM) \cite{Cen06}, Performance Factors Analysis (PFA) \cite{Pavlik09}, and the recency-weighted model by Gong and colleagues \cite{Gong11}. We then propose the Recent-Performance Factors Analysis (R-PFA) model. We evaluate this model's accuracy on a real-world dataset of student performance from the Assistments system \cite{Baker11}. Finally, since real-world datasets exhibit certain data limitations, we further examine the properties of the new R-PFA model and several alternatives on a range of simulated datasets.

\section{Prior Work in Performance Modeling}
To predict whether or not a student will succeed at completing a task, at a minimum, we ought to take into account some characteristic of the student and the task. There are two chief approaches to such modeling in the literature: graphical models, notably including Bayesian Knowledge Tracing, and logistic regression models.

In the best-known examples of logistic regression modeling, Item Response Theory and Rasch models include predictors relating to the ability of the student and the difficulty of the task. A refinement on this approach is Linear Logistic Test Models (LLTM) \cite{Fischer73,deBoeck04}. These logistic regression models replace the predictor relating to the difficulty of the individual tasks with a predictor that groups together tasks that share an underlying skill or Knowledge Component (KC). Because task difficulty is estimated from data, replacing per-task parameters with per-skill parameters reduces the number of model parameters, and leverages the power of task-level observations to provide a relatively more robust estimate of skill difficulty. 

The LLTM class of models includes Additive Factors Model (AFM) and Performance Factors Analysis (PFA) \cite{Cen06,Pavlik09,Chi11}. These models differ only in how they reflect prior practice to predict a student's future performance. The original LLTM is meant to reflect student knowledge during a short examination where we assume no learning is occurring, and therefore it does not include any summaries for the effects of practice, only effects of student and skill (i.e., student and KC intercepts). AFM introduces a slope coefficient for the total number of prior opportunities a student has had to practice a KC. The claim is that the more practice a student has had, the more likely they should be to get the next item correct. PFA decomposes the number of total prior practice opportunities into separate counts of successes and failures; with the assumption that successful and unsuccessful practice may have differential value for student learning and thus for probability of correctness on the next task.

\section{Model Comparison and Model Design}
Because performance modeling is rooted in statistics and machine learning, models of performance are often evaluated in terms of predictive accuracy. We take the position that models of task performance need to be interpretable above all. This stance disfavors models with good predictive accuracy when such models are black boxes, because such models make it difficult to advance the science of learning, or to develop systems that act on model predictions. Nonetheless, predictive accuracy is a sensible way to choose among multiple interpretable models.

We consider interpretability in terms of model realism and complexity. Model realism considers that many aspects of the world may affect student performance; a model should reflect as many of the biggest effects as possible, and do so as accurately or plausibly as possible in terms of both structure and estimated parameters. The danger of realistic models is that they can be highly complex. Models that are excessively complex may not be fully identified, or they may "overfit" the available data, i.e., they may reflect data characteristics that are minor at best, or inaccurate at worst.

By way of example, we can place Bayesian Knowledge Tracing (BKT) \& PFA on the realism-complexity continuum. BKT has a generative structure that represents (to a degree) human learning, but this structure also leads to mathematical complexity and may lead to implausible parameter values \cite{beck_identifiability:_2007}. PFA is relatively simpler mathematically because of its linear structure, but may still yield implausible parameters. For example, unless the parameters are artificially restricted, PFA may estimate that practice on a skill is associated with a decrease in the probability that a student will correctly answer a problem on that skill. BKT and PFA have comparable (and mediocre) predictive accuracy. \cite{Pavlik09,Gong11}

Accordingly, we aim for a model that is realistic, not excessively complex, and with good predictive accuracy. This is no small goal; for instance, models may have similar predictive accuracy but different parameter interpretations. For example, \cite{kaser_different_2014} find that AFM only estimates positive slopes for practice on about 50\% of the skills whereas other models estimate positive slopes for practice of almost all skills. However, if improvement in parameter plausibility does not lead to reliable improvement in predictive accuracy \cite{kaser_different_2014}, it is hard to decide which model is preferable.

We use AIC as an operational definition of model quality. AIC is a likelihood-based measure of model accuracy that incorporates a penalty for model complexity. Minimizing AIC is equivalent to minimizing KL-divergence risk.
An alternative technique for model comparison is cross-validation. An especially important technique in educational data mining is student-stratified cross-validation, where a model is trained on one set of students, and used to make predictions for a held-out set of students. In this way, one can claim to have a reasonable expectation of how well the model will perform on entirely new students. Still, AIC is known to be asymptotically equivalent to cross-validation \cite{Akaike85,Wasserman04,James13}. In fact, in section \ref{sec:sim} we demonstrate that AIC is superior as a measure of model fit to the oft-used cross-validation with a 0-1 loss function.

One way to consider the distinction between AIC and cross-validation is to consider that these measures represent different loss functions. Cross-validation can use any loss function, but 0-1 loss is most common, while the KL-divergence that AIC uses is more similar to an $L_1$ prediction error (PE) loss.

Let $L$ be the loss function, let $Y_{ij}$ the actual correct (1) or incorrect (0) outcome for student $i$ on a practice opportunity on skill $j$, 
and let  $\hat p_{ij} = \hat P(Y_{ij} = 1)$ be the estimated probability of a correct answer (i.e. the continuous output from the logistic regression model). 

\begin{eqnarray}
\text{0-1 loss} \label{eq:0-1loss}&~& L(\hat p_{ij}, Y_{ij}) = \left\{ 
                        \begin{array}{ll}
                        0 & \text{if ~} |\hat p_{ij} -Y_{ij}| < 0.5\\
                        1 & \text{otherwise}
                        \end{array}
                        \right. \\[10pt]
\text{PE loss} \label{eq:PEloss} &~& L(\hat p_{ij}, Y_{ij}) = |\hat p_{ij} - Y_{ij}| 
\end{eqnarray}

The primary difference between these loss functions is in whether we are interested in only prediction accuracy, or in accuracy and model confidence. The more confident a particular model is in it's predictions, the closer the estimated probability of a correct response, $\hat p_{ij}$, will be to the actual student response. 
Under PE loss, a hypothetical model that is accurate but not confident in its prediction is considered to perform worse than a model that is accurate and confident. By contrast, 0-1 loss discards information on model confidence, and treats confident and non-confident models equally. When $\hat p_{ij}$ is near 0.5, the two measures may disagree. The 0-1 loss function may prefer a model that has high predictive accuracy even if $\hat p$ is near 0.5, i.e., even if the model is not confident in its predictions. 
As model confidence increases, agreement in model ranking between the two loss functions will increase. 

For example, suppose that for a particular individual with two opportunities for practice on KC $j$, we observe $Y_{ij} = (0,1)$. Now suppose that Model 1 estimates the probability of a correct response as $\hat p = (0.48, 0.52)$, while Model 2 estimates the probability of a correct response as $\hat p = (0.1, 0.9)$. Then 0-1 loss will not distinguish between these two models, but PE loss will prefer the model that is more confident in predicting that the first attempt response is incorrect, and the second attempt a correct.

When cross-validation uses 0-1 loss, it ignores model confidence, but AIC considers both model accuracy and model confidence. The significance of this distinction will become apparent in the model comparisons below.

\subsection{Recent-Performance Factors Analysis}
Recent-performance Factors Analysis (R-PFA) focuses on recent history, rather than the complete practice history of a student \cite{galyardt_recent-performance_2014}. The first intuition behind this model is that having learned a skill makes it more likely that the student will get the next item correct; not having learned the skill makes an incorrect response more likely than a correct. The second intuition is that recent practice history with a KC may contain all the necessary information about whether or not a student has acquired the KC. We can relate this idea to a `moment-of-learning'. If a student has been successful with recent practice, then a moment-of-learning has likely already occurred. If recent attempts have not been successful, then the student has most likely not yet learned the KC. 

\subsection{Formal Model Descriptions}
To evaluate R-PFA comprehensively, we examine a number of alternative models. The notation we use differs from other publications of some models, but we hope that our consistent use of notation across all models will facilitate the comparison. We use the following notation:\\[4pt]
\begin{tabular}{ll}
 $j$ &  KC index, $j = 1, \ldots, J$\\
 $i$ &  student index, $i = 1, \ldots, N$\\
 $t$ & practice opportunity index, $t = 1, \ldots, O_{ij}$\\
  $X_{ijt}$ &  response by student $i$, on opportunity $t$ of KC $j$, \\
    	& $ X_{ijt} =  \left\{\begin{array}{ll} 
     		0 & \text{if  incorrect} \\
      		1 & \text{if correct} \\
   \end{array}\right. $\\
  $p_{ijt}$ & Probability of a correct response: $Pr(X_{ijt}=1)$ \\
  $T_{ijt}$ &  count of past opportunities\\
  $S_{ijt}$ &  recency-weighted count of previous successes, up to trial $t$\\
  $F_{ijt}$ &  recency-weighted count of previous failures, up to trial $t$ \\
  $R_{ijt}$ &  recency-weighted proportion of past successes \\
\end{tabular}\\[5pt]

All the models we examine are logistic regressions, where the general form is 
\begin{equation}p_{ijt} = Pr(X_{ijt}=1 | Z=z) = \frac{\exp(z^\prime \beta)}{1 + \exp(z^\prime \beta)}. \label{eq:logreg}\end{equation}
Each of the main models that we examine uses a different representation of a student's prior practice. These terms, which replace the generic $Z$'s in equation \ref{eq:logreg}, are displayed for clear comparison in table \ref{tab:terms}. The previously published models are Additive Factors Model (AFM) \cite{Cen07,Cen08}, Performance Factors Analysis (PFA) \cite{Pavlik09}, and PFA-decay \cite{Gong11}. We additionally include baseline models S-only, R-only, and R-AFM.

\begin{table}[tbp]
   \centering
      \caption{Terms in predictive model variants. }
   \begin{tabular}{@{} lcccccc @{}} 
      \toprule
       & Student & KC & Success & Failure & Total & Recent  \\
       & ability & difficulty & count & count & trials & success rate \\
      \midrule
      AFM    & $\theta_i$ & $\beta_j$ & & & $\gamma_j T_{ijt}$ \\   
      PFA    & $\theta_i$ & $\beta_j$ & $\alpha_j S_{ijt}$ & $\rho_j F_{ijt}$   \\[3pt]
      S-only & $\theta_i$ & $\beta_j$ & $\alpha_j S_{ijt}$ &  \\[3pt]
      R-only & $\theta_i$ & $\beta_j$ & & & & $\delta_j R_{ijt}$  \\[3pt]
      R-AFM  & $\theta_i$ & $\beta_j$ & & & $\gamma_j T_{ijt}$ & $\delta_j R_{ijt}$   \\[3pt]
      R-PFA  & $\theta_i$ & $\beta_j$ & & $\rho_j F_{ijt}$ & & $\delta_j R_{ijt}$  \\[3pt]
      \bottomrule
   \end{tabular}
   \label{tab:terms}
\end{table}

AFM represents prior practice as the total number of prior opportunities for a student to practice the KC: 
\begin{equation}
logit(p_{ijt}) = \theta_i + \beta_j + \gamma_j T_{ijt}. \label{eq:afm}
\end{equation}

PFA distinguishes effects of prior successes and prior failures in predicting future success:
\begin{equation}
logit(p_{ijt}) = \theta_i + \beta_j + \alpha_j S_{ijt} + \rho_j F_{ijt}. \label{eq:pfa}
\end{equation}

PFA-decay \cite{Gong11} is an adjustment to PFA that uses a decay weight to account for recency of observations:
\begin{eqnarray}
S_{ijt} &=& \sum_{p=1}^{t-1} d^{t-1-p}X_{ijp} \label{s-prime}\\[4pt]
F_{ijt}&=& \sum_{p=1}^{t-1} d^{t-1-p}(X_{ijp}-1) \label{f-prime}
\end{eqnarray}
Aside from the decay weight, PFA-decay uses the same predictors $S$ and $F$ as original PFA. In fact, when $d=1$, PFA-decay and PFA are exactly the same. Thus, we refer to both these models that only include (possibly decayed) counts $S$ and $F$ as PFA.

Another common approach in general regression modeling, is to perform a logarithmic transformation on count variables. In educational applications (e.g., \citeNP{yudelson14-javamooc,Chi11}), a logarithmic transformation of practice counts is an argument that practice beyond some threshold amount has only a marginal effect on the probability of a correct response. The transformation represents a sensible, realistic intuition about performance, but the regression coefficient on a log-transformed count is difficult to interpret. Moreover, the logarithmic transformation is simply a down-weighting of the total amount of practice, it does not account for any recency effects.

As an alternative for the count of successes, we introduce an exponentially decayed proportion of successes, $R_{ijt}$. 
\begin{equation} \label{eq:R}
R_{ijt} = \frac{\sum_{p=-2}^{t-1} d^{(t-p)}X_{ijp}}{\sum_{p=-2}^{t-1} d^{(t-p)}}
\end{equation} 
Aside from the decay weighting, which is explained below, the proportion of successes is quite simply the count of prior successes divided by the count of total prior attempts.

There are two issues to consider in decay weighting: the weighting function (kernel) and the weight strength. In non-parametric methods, the choice of kernel is generally less important than the decay weight \cite{Wasserman06nonpar}. An example of an alternate weighting function is the box kernel, where the tuning parameter is window size $k$ in the sense of the `last $k$ attempts'. The interpretation is simple, but box kernel treats all attempts within $k$ as equally important, which may not be sensible. In the PFA model, $k$ covers the entire practice history, which weighs all attempts as equally important, and does not discard even the oldest evidence. 

R-PFA and PFA-decay \cite{Gong11} both place an exponential decay weight on prior practice. Importantly, Gong et al. \citeyear{Gong11} fix $d$ at 0.9, aiming not to ``eliminate the effects of further practices too quickly.'' This is an overly simplistic choice, and as we shall demonstrate in section \ref{sec:assist-results}, simply tuning the decay parameter in the PFA models appropriately produces large improvements in predictive accuracy. In exponential weighting, different values of the decay weight $d$ control the `smoothing' of $S_{ijt}$ (PFA-decay) and $R_{ijt}$ (R-PFA) over the history of practice. If $d = 1$, then a student's entire history of practice gets equal weight. Alternatively, if $d = 0.1$, then 90\% of the weight is on the single previous trial, and 9\% is placed on the 2nd most recent attempt, so that effectively only the last attempt is counted in the recent history. 
Choosing a weight $d$ is precisely analogous to choosing smoothing bandwidth in nonparametric statistics (e.g., \citeNP{Wasserman06nonpar}, figure 4.5). 
For exponential decay, the decay parameter $d$ ranges from 0 to 1, while for the box kernel, the window size $k$ ranges from 1 to infinity, so that selecting the optimal $d$ has a more tractable search space. Thus the exponential decay function has both a computational advantage for tuning decay weight, and interpretability advantage since older evidence is down-weighted. 

\begin{figure}[p]
\begin{center}
\includegraphics[width=0.85\linewidth]{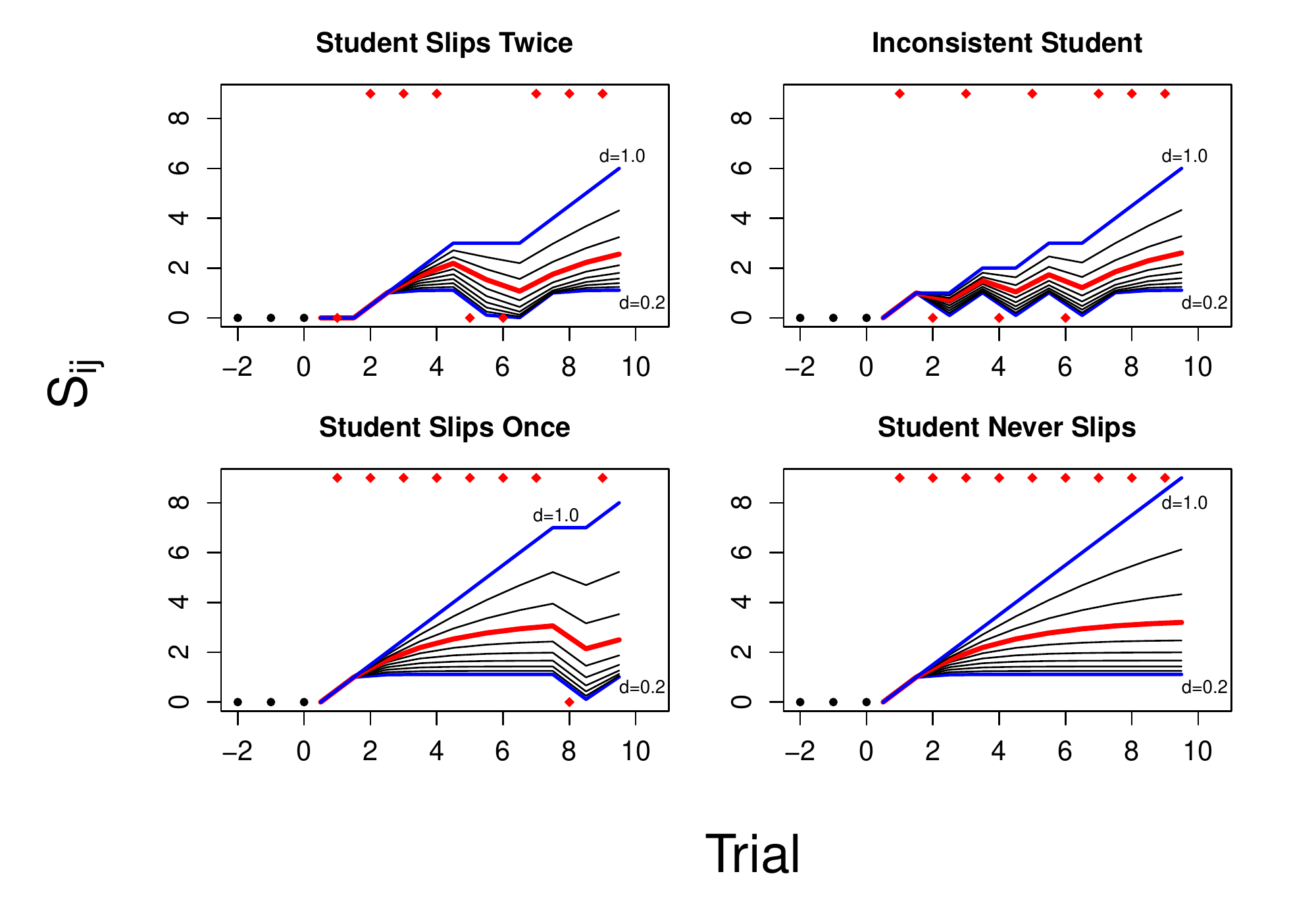}

\vspace{.5cm}

\includegraphics[width=0.85\linewidth]{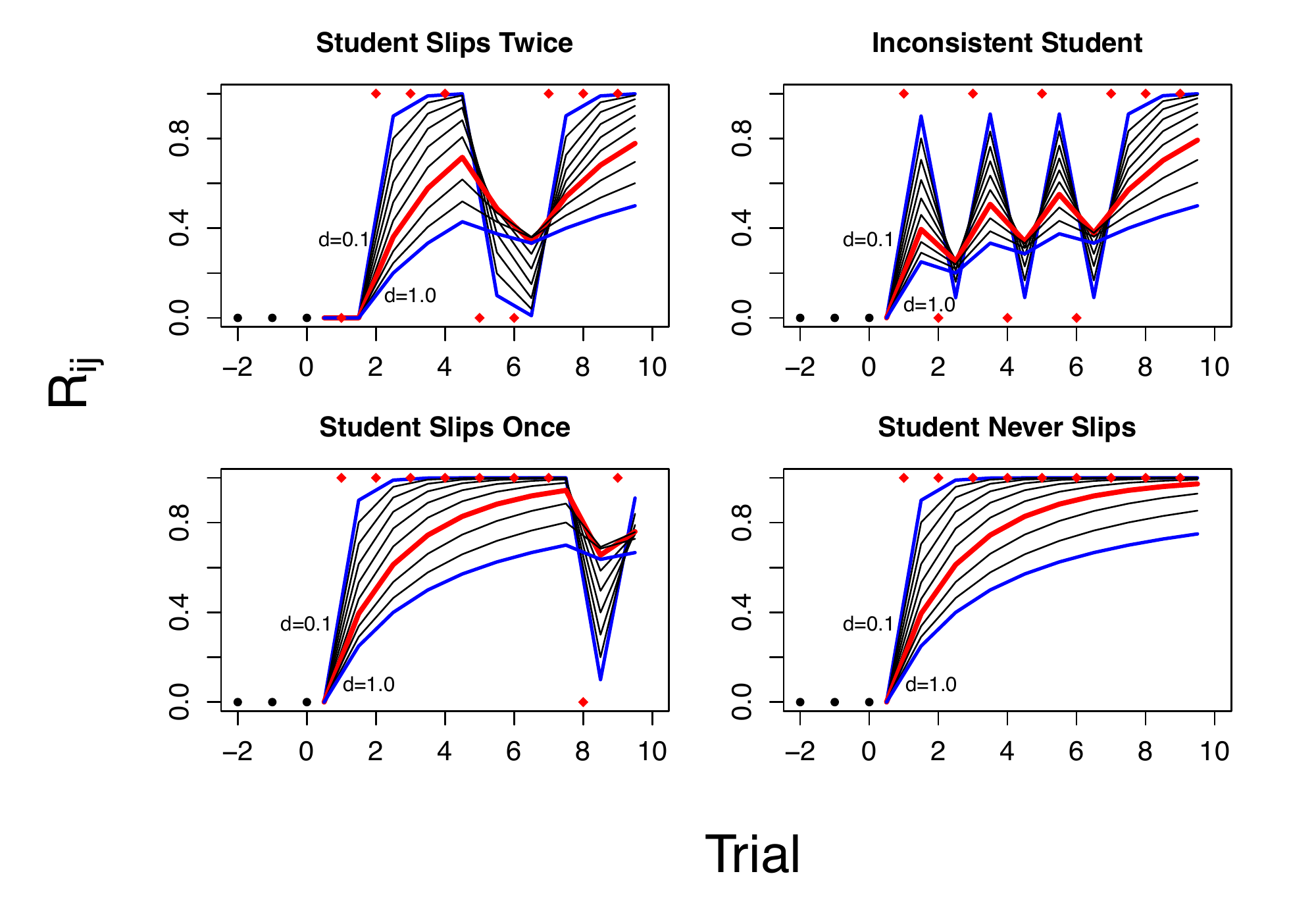}
\caption{Effect of exponential decay weighting on the count of successes $S_{ijt}$ and proportion of successes $R_{ijt}$ given distinct patterns of student behavior. Red diamonds signify correct and incorrect responses (at 1.0 and 0.0 on the vertical axis), and black circles signify the ``ghost" attempts. Each line indicates a different value of the decay parameter $d$, the thick red line indicating $d=0.7$.}
\label{Traceplot}
\end{center}
\end{figure}

We can see the effect of the different values of $d$ most clearly in the pattern ``Student Slips Twice" (Figure \ref{Traceplot}). This student has the attempt history $X_{ij} = (0,1,1,1,0,0,1,1,1)$, as indicated by the red diamonds in the figure. With a decay weight $d=0.2$, after 1 error followed by 3 corrects, $S_{ij5} = 1.24$, but then when the student misses the next item, $S_{ij6}$ drops to 0.248. Without decay, i.e., when $d=1.0$, $S_{ijt}$ grows slowly with each item that a student gets right, and never decreases after errors. For the highlighted decay weight $d=0.7$, $S_{ijt}$ increases at a moderate pace with each correct, until $S_{ij5} =2.19$. Then, when the student answers incorrectly on trial 5, $S_{ij6} =1.53$, a small drop, and with the subsequent error drops further to $S_{ij7} = 1.07$. The impact is similar on the proportion of successes $R_{ijt}$.

The behavior of $S_{ijt}$ and $R_{ijt}$ with $d=0.7$ mirrors our intuition. If we were tutoring a student one-on-one, on the third correct attempt in a row, we might think `Ok, they've mastered this skill.' When the next attempt is incorrect, we might think `That was probably just a slip.' On the second incorrect attempt, we might revise our assessment of the student's knowledge: `Hmm, maybe they don't know this.' But after 3 subsequent correct responses in a row, we might be fairly convinced the student has learned the KC. This parallels exactly the Bayesian updating of the probability that a student has learned a KC that takes place in a Bayesian Knowledge Tracing (BKT) model. In this way, exponential decay weighting is capturing student performance in a similar way BKT, but without the complexity of a Hidden Markov Model.

The recency-weighted {\em proportion} of successes $R$ is similar to the recency-weighted {\em count} of successes $S$, but there are differences. The interpretation of $R_{ijt}$ is consistent across different values of the decay parameter. If $R_{ijt}$ is near 1, then the student has been successful in recent attempts; if it is near zero, then the student has recently been unsuccessful. If the student has a fully successful history of practice, $R$ will converge to 1 no matter the value of the decay weight. By contrast, $S$ does not have a consistent interpretation. For any value of $d$, $R$ is scaled to fall between 0 and 1, implying that $R$ is easily interpretable as some proportion, e.g., `a student has a success rate of about 80\% over the last few items.' By contrast, $S$ has asymptotic properties that complicate interpretation. Since each $X_{ijt}$ is either 1 or 0, the counts $S_{ijt}$ and $F_{ijt}$ are bounded by they geometric series $\sum_{p=1}^\infty d^p = {(1-d)}^{-1}$. If $d=1$, the series does not converge. For every $d<1$, the series will converge to a different number. The asymptotic limit of this series is visible in  the pattern ``Student Never Slips'' in figure \ref{Traceplot}. For $d=0.9$ the limit is ${(1-0.9)}^{-1}= 10$; for $d=0.2$, the limit is 1.25. The meaning of a particular value of $S$, but not $R$, depends on $d$.

The consistent interpretation of $R$ also allows us to interpret $R$ as a proxy for whether or not a student has experienced a moment of learning. As an example, consider two students with histories $X_{ij} = \{0, 0, 1, 1, 1, 1\}$, and $X_{ij} = \{0, 0, 1, 1, 1, 1, 1, 1, 1, 1\}$. Intuitively, we would tend to believe that both of these students have experienced a moment of learning and are likely to get the next item correct. For any value of $d$, these two students will have a similar $R$ value. However, for a small value of $d$, $S$ will be the same for these two students, but for a $d$ closer to 1, $S$ will be different, and the predictions will be different. Thus $S$ is not interpretable as a proxy for a moment of learning. This gives an interpretative advantage to the recency-weighted proportion $R$, and may or may not give a predictive advantage as well.

Early observations of practice on a skill necessarily contain less evidence of student mastery than the accumulation of early and later observations. Thus, both $R_{ijt}$ and $S_{ijt}$ are noisy on early attempts on a KC. To illustrate, consider two students: the first student has the performance history of $X_{ij} = (1,0)$. The second student has a performance history of $X_{i^\prime j} = (0,1,1,1,1)$. We would be highly doubtful that the former student has mastered the KC, while the latter student has likely mastered the KC. After one trial, the proportion of recent successes is 1 and 0, respectively. The first student has a higher proportion of success after the first trial than the second student does after 5 trials, including 4 successful attempts in a row. Thus, the proportion of successes is a noisy representation for the first student.

To adjust for this noise on the first few attempts, $R_{ijt}$ incorporates the assumption that 3 attempts prior to the first attempt would have been incorrect. That is, we stipulate ghost attempts $X_{i,j,-2} = X_{i,j,-1} =X_{i,j,0} = 0$. This is making explicit an assumption that at time 0, a student has not already learned the KC, which is very plausible in educational data. These ghost attempts only affect the calculation of $R_{ijt}$, i.e., they do not affect $T_{ijt}$, and they are not extra instances in the dataset. Note that such ghost attempts implicitly also exist in the calculation of $S_{ijt}$, in the sense that on trial one, the count of previous successes is zero. The ghost attempts are included in equation \ref{eq:R} and figure \ref{Traceplot}. 

\paragraph{ Model variants including R}

To separate the effects of recent practice, total practice, and the differential predictive effects of recent success and failure, we compare three model variants that contain $R$: 
\begin{eqnarray}
\text{R-only} &~& logit(p_{ijt}) = \theta_i + \beta_j + \delta_jR_{ijt}, \label{eq:ronly} \\[3pt]
\text{R-AFM} &~& logit(p_{ijt}) = \theta_i + \beta_j + \gamma_jT_{ijt} + \delta_jR_{ijt}, \label{eq:rafm} \\[3pt]
\text{R-PFA} &~&  logit(p_{ijt}) = \theta_i + \beta_j +  \rho_j F_{ijt} + \delta_j R_{ijt}. \label{eq:rpfa}
\end{eqnarray}

We compare these three recent-history models with the established AFM and PFA models, as well as the S-only baseline model that uses only the count of successes. For PFA \& R-PFA, which include two decay-weighted variables, we consider both the case where the tuning parameters are equal and the case where they are tuned separately. This allows for the potentially differential predictive power of recent successes vs. recent failures.

\section{Model Application to Real-world Data}\label{sec:assist}
\subsection{Methods}

We evaluate the models described above in modeling student performance in the Assistments data used in the ``moment of learning'' work by Baker and colleagues \cite{Baker11}. The data contain first attempts by 4138 students on problem sets involving 54 knowledge components (KC), for a total of 187,309 first attempts. Each problem is coded with only a single KC. Each KC was attempted between 89 and 16,200 times, and had an overall percent correct between 23\% and 95\%. The data are from the mastery learning ``Skill Builder'' feature of Assistments, which allows teachers to set a threshold for the number of problems a student must correctly answer in a row to be considered proficient. For this data set, the threshold was set at either 3 or 5. 

This data set is sparse at the student level. First, the median number of KCs seen by each student is 3, and 75\% of students practice 7 or fewer different KCs. Second, the median number of total attempts per student summing across all KC's is 20, and 435 students (11\%) made 3 or fewer total problem-solving attempts. This sparsity of data at the student level means that any student effects in a model should be fit as random effects coming from a common distribution. In this way, we `pool' the data, so the student effects $\theta_i$ for students with less data shrink towards the mean student effect. The ghost attempts necessarily have the greatest influence on practice strings that are relatively short, i.e., they reduce the noise that would otherwise be present in $R_ij$ for these attempts. 

There are a large number of students per KC; of the 54 KC's, only one is practiced by fewer than 25 students, and the median number of students per KC is 410. However, the number of attempts for each student on each KC is small, with a median of 4, and a mean of 8. For this reason, we also treat all KC intercepts and slopes as random effects. We did not include the covariance matrix for KC parameters in the model.

The number of students per KC makes student-stratified cross-validation unreliable, if not entirely untenable. There are 54 KCs, but 27 of them are encountered by fewer than 410 students, i.e., fewer than 10\% of the students.  Moreover, the sparsity is not uniform; which KCs were attempted by particular students is not uniformly random. Omitting 10-20\% of the students in a cross-validation fold leads to omitting a number of KCs. Therefore 5-fold or 10-fold cross-validation will result in very poor estimates of KC parameters, or an inability to use the model to make predictions. Instead, AIC is used as the measure of model fit; as we demonstrate in section \ref{sec:sim}, AIC is at least as reliable, if not better than cross-validation. 

We used the {\tt glmer} function in the R package {\tt lme4} to fit all models listed in Table \ref{tab:terms} \cite{lme4}. Counting all of the different tunings of relevant decay weights, we fit a total of 111 models, though below we display only the results for the most illuminating comparisons. Data\footnote{\url{https://sites.google.com/site/assistmentsdata/home/goldstein-baker-heffernan}} and analysis code\footnote{\url{https://sites.google.com/site/aprilgalyardt/research}} are posted online.

\subsection{Results and Discussion}\label{sec:assist-results}
\begin{figure}[p]
\begin{center}
\includegraphics[scale=0.8]{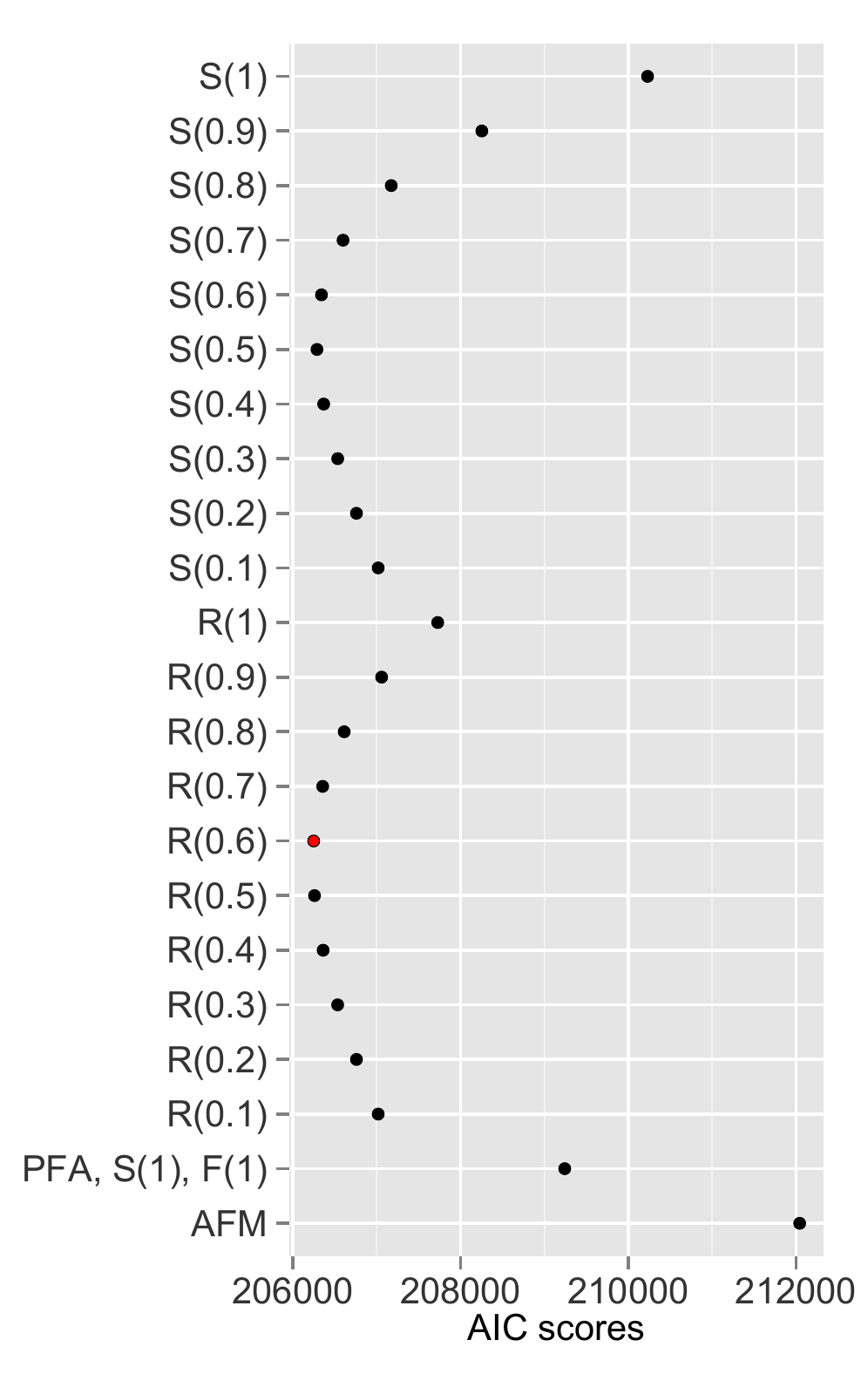}
\caption{AIC scores for S-only, R-only, PFA and AFM on Assistments data. Models are labeled with the decay parameter in parentheses. Smaller AIC is better. The best model in this set of models is R-only $d=0.6$, marked with a red triangle.}
\label{fig:aic1}
\end{center}
\end{figure}

\paragraph{Importance of Recency Weighting}
We first compare PFA and AFM to models where the prior practice representation is the recency-weighted count of successes $S_{ijt}$ (S-only) or the recency-weighted proportion of successes $R_{ijt}$ (R-only), as in figure \ref{fig:aic1}. First, we find that PFA outperforms AFM, replicating prior research \cite{Pavlik09,Chi11}. Second, S-only with decay weight $d=1$ outperforms AFM. With a decay parameter of 1, $S_{ijt}$ is simply the total count of all prior successes for person $i$ on KC $j$. Thus, a simple count of successes is a better predictor of future success than the total count of practice.

Third, recent success is a better predictor of learning than the entire history of practice, since both S-only and R-only outperform AFM and PFA. Fourth, $R_{ijt}$ and $S_{ijt}$ have the same predictive value when the decay parameters are small, $d < 0.3$, but $R$ becomes a more powerful predictor than $S$ as the decay parameter increases. In other words, as the predictor includes more practice history, the proportion of successes becomes more valuable than the count of successes. With $d=0.9$, AIC for R-only is 1200 less than AIC for S-only, a substantial difference.

\begin{figure}[p]
\begin{center}
\includegraphics[scale=0.88]{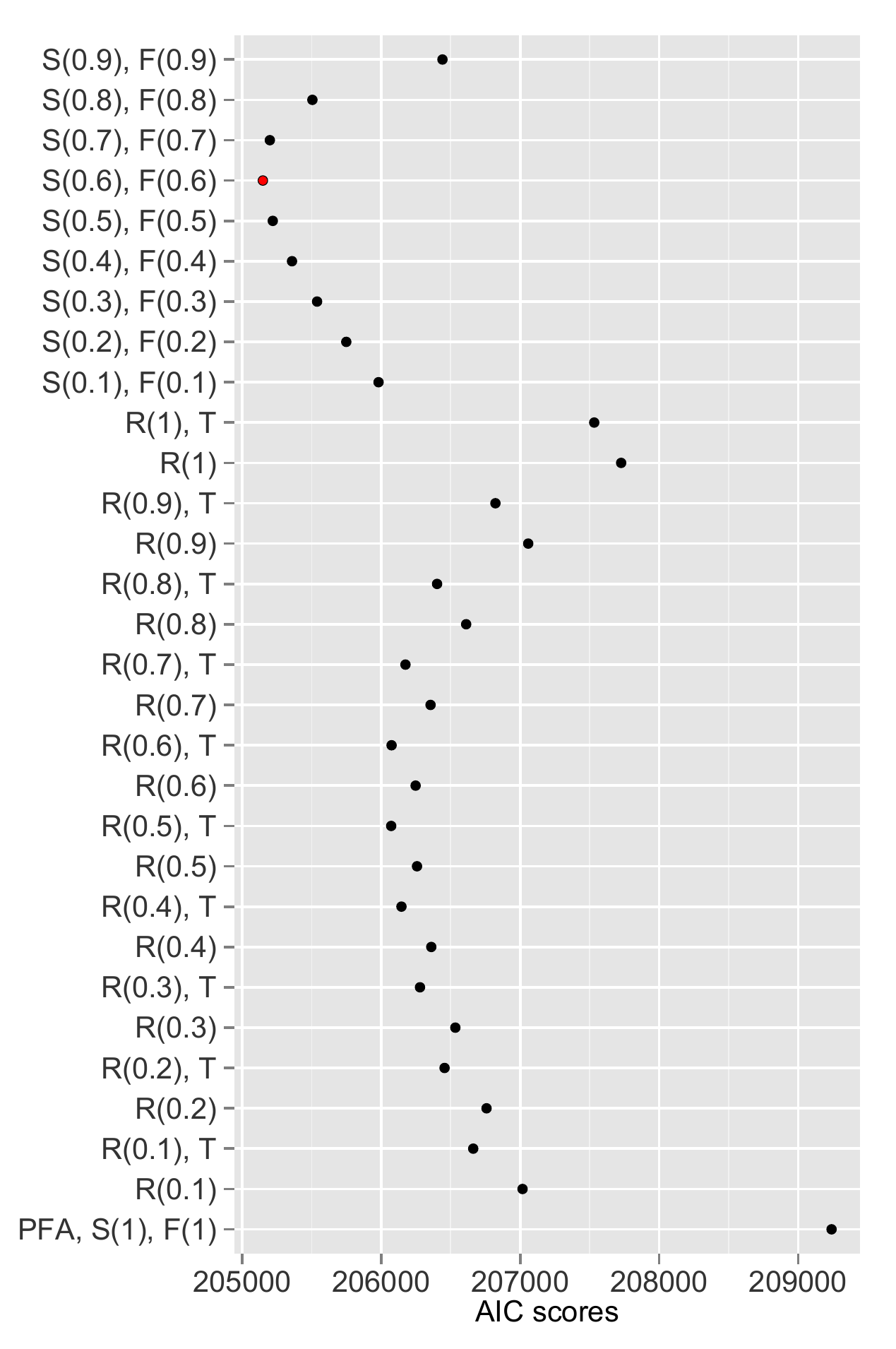}
\caption{AIC scores for R-AFM, R-only, PFA with equal S and F decays, and classic PFA on Assistments data. Models are labeled with the decay parameter in parentheses. Smaller AIC is better. The best model in this set of models is PFA $d=0.6$, marked with a red triangle.}
\label{fig:aic2}
\end{center}
\end{figure}

\paragraph{Importance of Failures and Total Practice}
Given the baseline value of tracking recency-weighted successes, which already outperforms PFA without recency weighting, what is the value of additionally incorporating total practice or failed practice? Comparing R-only and R-AFM enables us to judge the additional predictive value of amount of total practice compared to recent success rate, and recency-weighted PFA (holding constant the decay weight for the success and failure counts) allows us to judge the additional predictive value of failed practice (figure \ref{fig:aic2}). We find that adding a predictor for total amount of practice (R-AFM) improves on the performance of R-only, but recency-weighted PFA with separate success and recent failure counts produces even larger gains in predictive accuracy. The best model so far is PFA with $d=0.6$, i.e., that the most relevant information for predicting future performance is contained in the most recent 3-5 attempts, including separate counts of successes and failures.

\begin{figure}[p]
\begin{center}
\includegraphics[scale=0.9]{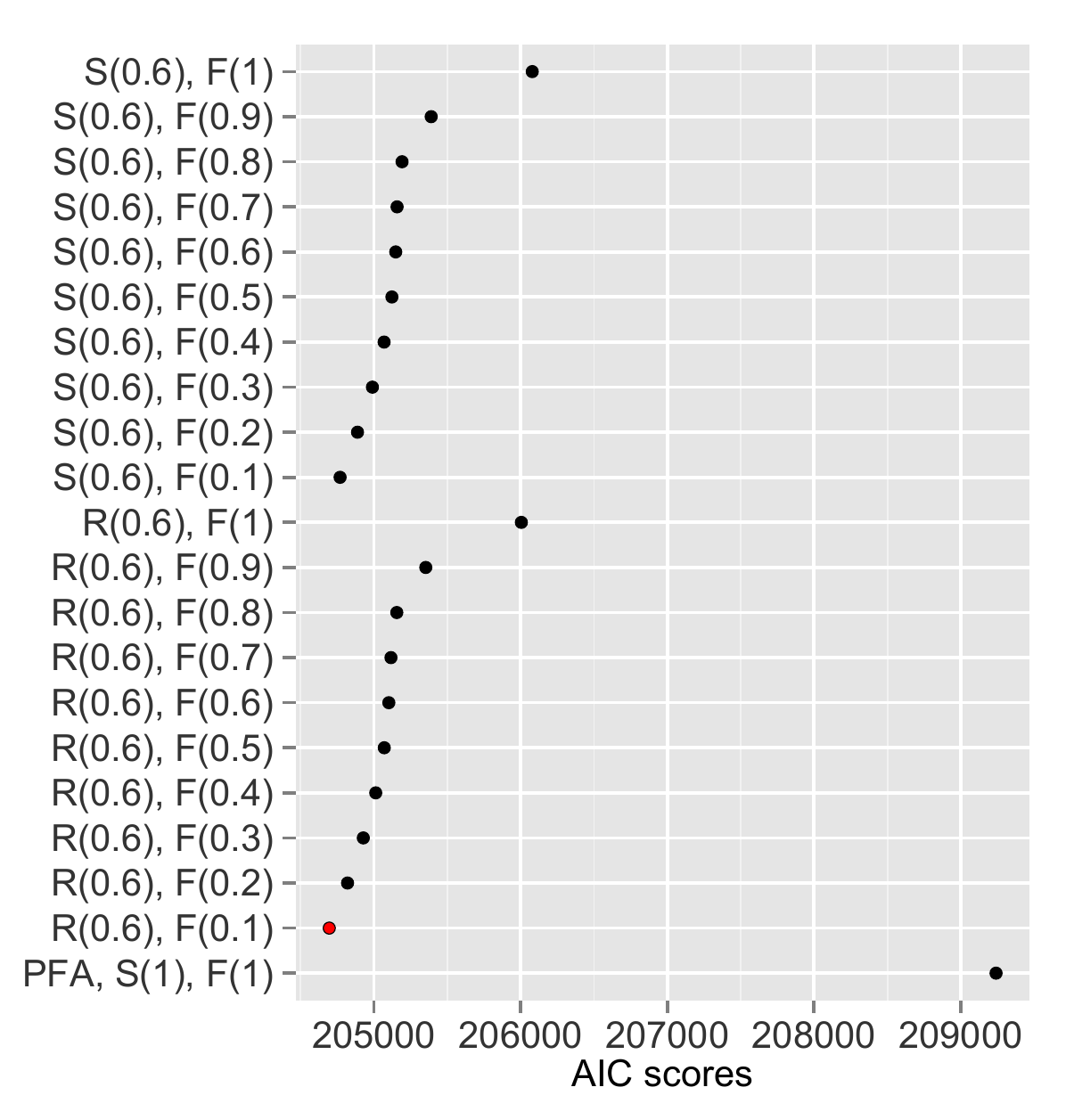}
\caption{AIC scores with differential success and failure decays on Assistments data. For success, best-so-far decay is $d=0.6$ for $R$ in R-PFA, and also $d=0.6$ for $S$ in PFA. Models are labeled with the decay parameter in parentheses. Smaller AIC is better. The best model in this set of models is R-PFA with success decay $0.6$ and failure decay $0.1$, marked with a red triangle.}
\label{fig:aic3}
\end{center}
\end{figure}

\paragraph{Comparing Count {\em versus} Proportion of Successes}
As seen in figure \ref{fig:aic1}, $R$ alone is a better predictor than $S$ alone once $d$ grows sufficiently large, even though they contain similar information. This finding stands even after incorporating failure information (figure \ref{fig:aic3}). At each value of the decay parameter, R-PFA (with $R$ and $F$) outperforms PFA ($S$ and $F$), and once again, the difference increases as the recency weight approaches 1.

\paragraph{Differential Recency of Successes and Failures}
Starting with the PFA model, we allow the decay weight for $F$ to vary, while holding constant the decay weight for $S$ at $d=0.6$, which was optimal in the R-only, S-only, and PFA models (figure \ref{fig:aic3}). We find that failure counts deserve the lowest possible decay weight, implying that only a failure on the single most recent attempt contains relevant information for predicting future performance, and prior failures are less informative. The result is the same for the R-PFA model, allowing the decay weight for $F$ to vary, while holding constant the decay weight for $R$ at $d=0.6$.

\begin{figure}[p]
\begin{center}
\includegraphics[scale=0.92]{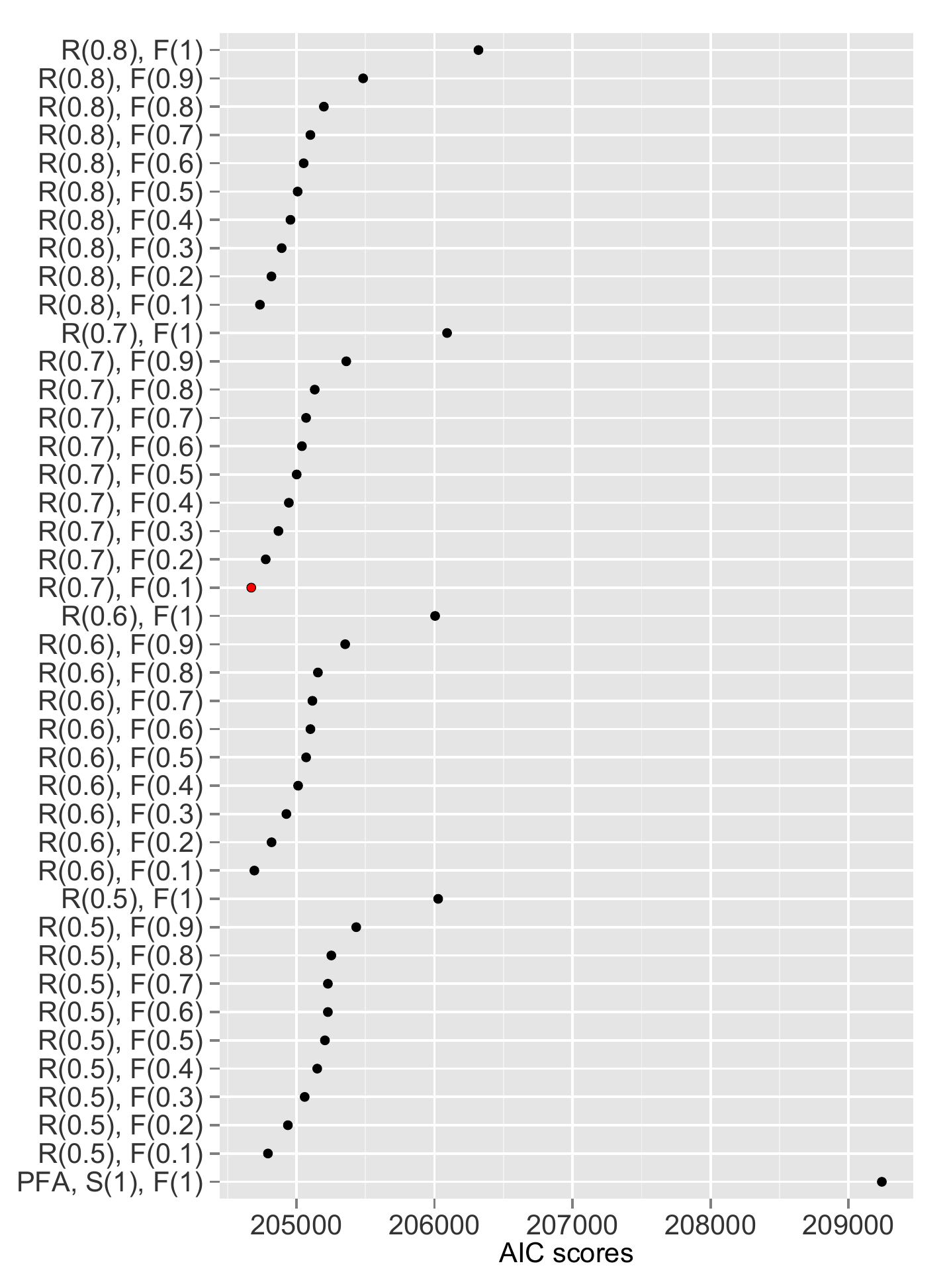}
\caption{AIC scores for Assistments data. Models are labeled with the included predictors and the decay parameter in parentheses. Smaller AIC is better, the best model in this set of models is R(0.7), F(0.1), marked with a red triangle. This is the best-performing model overall.}
\label{fig:aic4}
\end{center}
\end{figure}

Figure \ref{fig:aic3} suggests that tuning the success rate and the failure rate separately offers a distinct advantage. The final set of model comparisons verifies this finding. Figure \ref{fig:aic4} shows R-PFA models where the recency weight for $R$ varies between 0.5 and 0.8, and the recency weight for $F$ ranges from 0.1 to 1.0. (Only the R-PFA models are shown, since the PFA models using $S$ performed uniformly worse than equivalent models using $R$.) 

Of the models we compared on this dataset, the model with the highest predictive accuracy is R-PFA with recency weight $d=0.7$ for the proportion of successes $R$, and recency weight $d=0.1$ for count of failures $F$. For successes, with $d = 0.7$, the weights of the last 6 actions are respectively: $\{0.340, 0.238, 0.167, 0.117, 0.082, 0.057\}$. The 5 most-recent actions receive substantial weight, but 58\% of the weight is on the two most-recent actions. In contrast, with $d=0.1$, the weights for failures on the last 6 actions are:$\{ 0.9, 0.09, 0.009, 0.0009, 0.00009, 0.000009\}$. Applying the weight, if the last action is incorrect $d\cdot F \approx 0.9$, and if the last action is correct, $d\cdot F\approx0.1$. Thus, in this best-performing model, $R$ acts like a running average over the last 2-5 actions, while $F$ is effectively a binary indicator for whether the last action was correct or incorrect. 

The difference between the optimal tuning parameters for recent successes and recent failures may also be accounting for the difference in slips and guesses. If a student knows the KC and has been correctly responding, then $R \approx 1$ and $F \approx 0$. If this student then slips and responds incorrectly, with the optimal decay parameters, $R$ will decrease to 0.7, and $F$ jumps to 0.9. If the incorrect answer was truly a slip then the student will likely answer correctly on the next attempt, so that $R$ increases towards 1 again, and $F$ falls back toward zero. (See also Figure \ref{Traceplot}, ``Student slips once".)   In this way, $R$ is largely unaffected by slips, while $F$ is an indicator that the last response may have been a slip. Now consider a student who does not know the KC, and has a history of incorrect practice attempts, so that $R \approx 0$ and $F \approx 1$. If this student then guesses correctly on an item, $R$ only increases to 0.3, and $F$ falls to 0.1. Here $R$ is largely unaffected by the correct guess, while $F$ is an indicator that the last response may have been a guess. 

\subsection{Interpreting the Best-Performing Model}
The best overall model for predicting future success from a student's history is R-PFA with 3 parameters for each KC: $\beta_j$, the `easiness' of the KC; $\rho_j$, the effect of recent failures with the KC, and $\delta_j$, the effect of recent successes with the KC. To examine model parameters in detail, consider that in a logistic regression model with random effects, the estimates of the coefficients may not be normally distributed when the data is sparse. This means that it may be inappropriate to use the estimated standard errors to obtain confidence intervals for the parameters. To address this issue, we re-fit the best-performing model using a Markov Chain Monte Carlo algorithm, as implemented in the {\tt MCMCglmm} package in R \cite{MCMCglmm}. 

\begin{figure}[htp]
\begin{center}
\includegraphics[width=\linewidth]{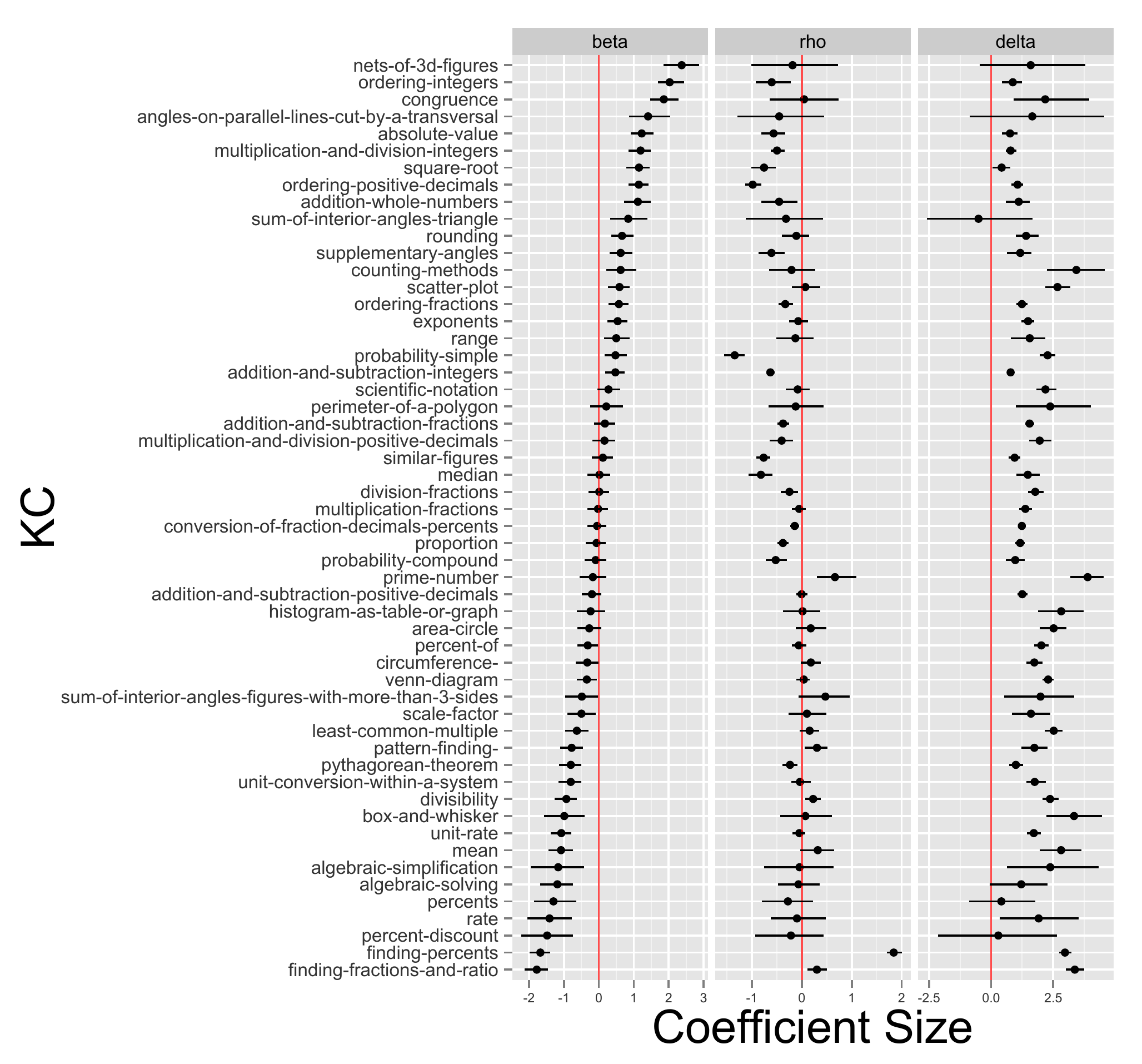}
\caption{Estimates for KC parameters in the best performing model: $logit(p_{ijt}) = \theta_i + \beta_j + \rho_j F_{ijt} + \delta_j R_{ijt}$. The decay weight for $F$ is 0.1, and for $R$ is 0.7. For each KC, the dot indicates the posterior median for the parameter and the line indicates the 95\% credible interval. KC's are ordered by their $\beta$ estimate with easier KCs at the top. }
\label{fig:coef}
\end{center}
\end{figure}

We examine the 95\% posterior credible intervals (CI) for each KC parameter (Figure \ref{fig:coef}). Recall that these were estimated with no restrictions on any coefficients. First, for 49 of the 54 KC's, the $\delta$ coefficients for the effect of recent successes are significantly positive. The remaining 5 KCs have very wide CI's and are not significantly different than zero. These 5 KCs are among the easiest and hardest KCs, and were practiced by few students. In general, the more recent successes a student has had, the higher the probability of correctly responding to the next item, which corresponds to our intuitions about learning. 

We further examined the covariance among the KC parameters. The 95\% CIs are $r(\beta, \rho) = (-0.62, -0.36)$, $r(\rho, \delta) = (0.32, 0.60)$, and $r(\beta, \delta) = (-0.49, -0.10)$. Notably, there is a significant negative correlation between KC easiness and the effect of recent failures; for relatively more difficult KCs, the effect of recent failure on predicting a correct response is positive, while for relatively easier KCs, the effect of recent failure is negative. With easy KCs, recent failure would predict subsequent failure for students who are not acquiring the KC, or who are engaged in non-productive behaviors, e.g., gaming the system. Interestingly, for difficult KCs, recent failure is positively associated with subsequent success.

It has been previously documented in PFA and AFM that the slopes for the effect of the count of past failures $F_{ij}$ (and occasionally even for the count of past successes $S_{ij}$) are often negative, e.g., \cite{kaser_different_2014}. Such negative slopes signal an area of concern (with the performance model itself or with the KC decomposition), because more practice, successful or unsuccessful, should increase the probability of a correct response. The R-PFA result that recent success is predictive of future success counters the negative-slope phenomenon.

   

\begin{sidewaysfigure}[p]
\begin{center}
\includegraphics[scale=0.7]{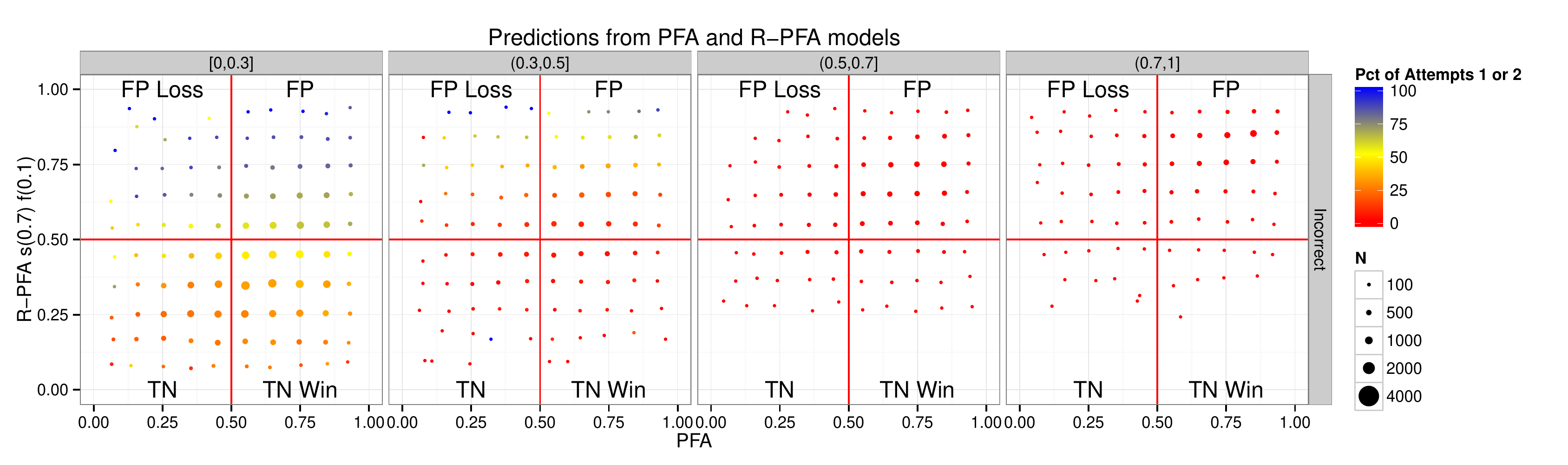}
\includegraphics[scale=0.7]{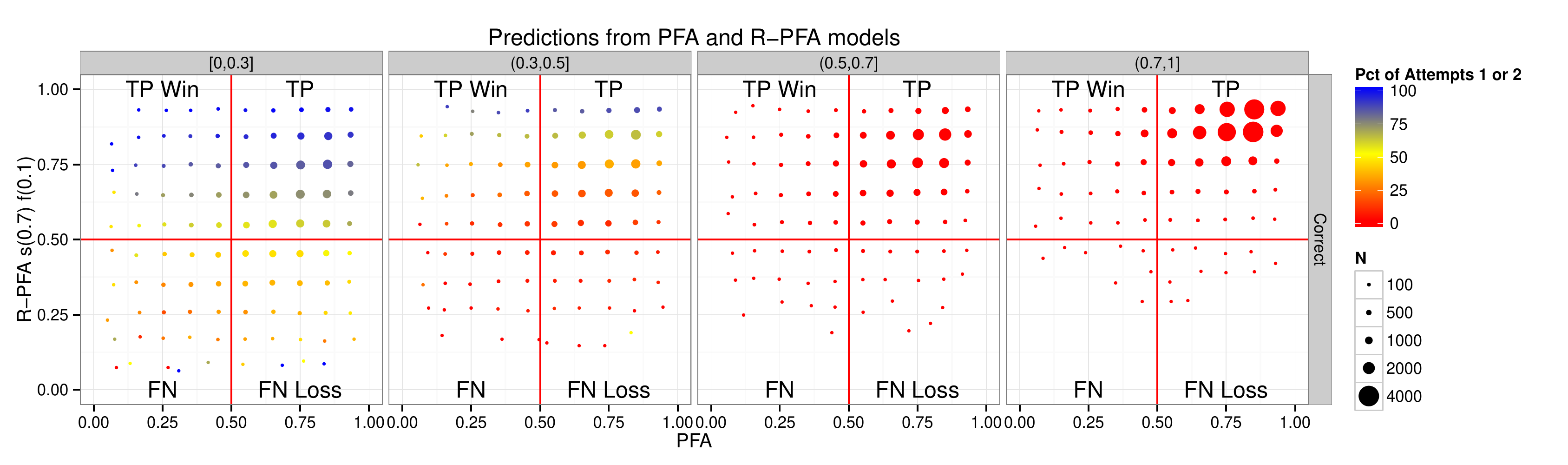}
\caption{PFA vs. R-PFA predictions in terms of the $\hat p$ value from the logistic model. The bubbles display $\hat p$ values from the two logistic models; the more data points there are near that $(x,y)$ position, the bigger the bubble. The color of the bubble indicates the percentage attempts in that position that are 1st or 2nd attempts by a student on a particular skill. The two rows of the figure correspond to actually incorrect (top) and actually correct (bottom) outcomes. Each row is divided into 4 facets according to the value of the $R$ predictor. The closer $\hat p$ is to the boundary values of 1.0 or 0.0, the more confident the model is in the prediction. The abbreviations are: TP - true positive, FP - false positive, TN - true negative, FN - false negative. Win and loss are for R-PFA relative to PFA A $\hat p > 0.5$ is a prediction of a correct outcome, and $\hat p < 0.5$ is a prediction of an incorrect outcome. }
\label{fig:phats}
\end{center}
\end{sidewaysfigure}

\paragraph{What is the source of R-PFA's advantage over PFA?} Although it is inappropriate to examine errors in prediction on a held-out or cross-validation set given the sparsity in our dataset, even comparing predictive accuracy on the training set is very illuminating. We present the difference in the predictions in figure \ref{fig:phats}. The two rows of the figure correspond to actually incorrect (top) and actually correct (bottom) outcomes. Each row is divided into 4 facets according to the value of the $R$ predictor: 

\begin{itemize}
\item $R$ in $[0,0.3]$ indicates that the student has produced either 1 or fewer right answers in the last 4 attempts, or is at the very beginning of practice.
\item $R$ in $(0.3,0.5]$ indicates 2 correct answers in the last 3-4 attempts.
\item $R$ in $(0.5, 0.7]$ implies that the most recent 2 answers were correct.
\item $R$ in $(0.7, 1]$ means that at least the last 3 answers were correct.
\end{itemize}

 The X and Y axes indicate the predictions from the PFA and R-PFA, respectively. This $(x,y)$ position has a different meaning for the actually correct and actually incorrect outcomes. For example, the top-right quadrant for the actually incorrect outcomes indicates false positive values, due to both PFA and R-PFA wrongly predicting that the student will respond correctly. The top-right quadrant for the actually correct outcomes indicates indicates true positive values, due to both PFA and R-PFA accurately predicting that the student will respond correctly. 

There are notable difference between the models in two cases, roughly corresponding to very early practice on a skill and to relatively late practice. First, when the true outcome is an incorrect response and the student has had few recent successes, R-PFA is much better at predicting these incorrect outcomes than PFA (top row, $R$ in $[0,0.3]$, TN Win). This is most often the case when the attempt is after the second (note the bubble color); both PFA and R-PFA often wrongly predict a correct response for this $R$ value when the attempt is the first or the second (top row, $R$ in $[0,0.3]$, FP). This improvement in predicting when a student will fail to answer correctly is an important contribution of R-PFA for intelligent tutors and adaptive systems. 

Second, when the student has had successes on the most recent items, R-PFA is more likely to predict a correct outcome than PFA. This is true both when the true outcome is correct, and when it is not, i.e., when the incorrect outcome is likely a slip. Ultimately, the number of false positive losses for R-PFA (top row, $R$ in $(0.5,0.7]$ or $(0.7,1.0]$) is much lower than the number of true positive wins (bottom row, same $R$). To an intelligent tutor, accurately predicting slips is arguably unimportant. An intelligent tutor using R-PFA rather than PFA would be more aggressive and more accurate at predicting student mastery of a skill, allowing students to graduate from practicing a skill more quickly than PFA.

When the student has had 2 correct answers in the last 3-4 attempts ($R$ in $(0.3,0.5]$), it is hard to know whether to expect a correct or an incorrect outcome. In the aggregate, PFA and R-PFA perform comparably in this case.

\section{Simulation Study} \label{sec:sim}
The purpose of the simulation study was to compare R-PFA against other performance models without the limitations of real datasets. Two characteristics of the Assistments dataset examined above complicate a thorough model comparison. First, the data sparsity in the Assistments data (section \ref{sec:assist}) precludes the use of cross-validation for model ranking. Nonetheless, cross-validation is a popular tool model comparison because holding out data during the model training can help prevent overfitting. Thus, we compare R-PFA to the other models both according to cross-validation and according to AIC. By using simulated data, we demonstrate that AIC is not only an appropriate model selection measure, but that it is {\em better} than cross-validation with the oft-used 0-1 loss function, and it can accommodate sparse data.

Second, the stopping criterion used in the Skill Builder feature of Assistments leads to data missing non-randomly. Once Assistments determines that a student has mastered a skill, there are no further practice opportunities for the student on this skill. In fact, we expect that no mastery criterion is perfect, and even ``mastered' students may have future incorrect practice. Thus, we would like to use data with evidence of post-mastery performance to train and evaluate our models. We demonstrate that decay weight $d$ in the 0.6-0.8 range is optimal even when there is no stopping rule that affects data generation.

\subsection{Methods}
We first simulated data from the Bayesian Knowledge Tracing (BKT) model and two adaptations of BKT. We then compared the fit of seven logistic test models on each simulated data set. Classic BKT describes `ideal' student behavior, which may not capture all student behavior. Our two adaptations of the BKT model address this by incorporating more realistic student behavior. Thus, we can compare the logistic models in the presence of less than ideal student responses.  

In one adaptation of the classic 2-state BKT model, we posit a {\em 3-state BKT} model. In classic BKT, there are two states: a learned state where the student has a high probability of correctly responding to a question, and an unlearned state where the student has a low probability of correctly responding. In the {\em 3-state BKT} model, the states are unlearned, practicing, and fluent. In the unlearned state, students have a very low probability of correctly responding to a question. In the fluent state, students have a very high probability of correctly responding. In the practicing state, students are learning the KC, but their understanding is not complete, so they have only moderate probabilities of a correct response. When generating data, this specification will produce more interwoven sequences of 0's and 1's than the 2-state BKT model. For example, we will see more patterns like $X_{ij} = (0,0,1,0,1,0,1,1,0,1,1)$, rather than primarily patterns of the form $X_{ij} = (0,0,0,0,1,1,1,1)$. In this way, 3-state BKT incorporates realistic ``struggling" students' behavior. This 3-state model only serves a generative purpose; data generated from the 3-state model could also be fit by the 2-state model.

In the {\em (BKT+FS)} adaptation to the BKT model, we vary the behavior of different simulated students. We include a small proportion of students who occasionally engage in unproductive learning behavior that produces long strings of incorrect responses. In real datasets, such data may be produced by various causal mechanisms, e.g., by lacking mastery of a prerequisite KC, by abusing hints \cite{aleven_limitations_2000}, or by gaming the system \cite{baker_off-task_2004}. Specifically, a fraction of students may be likely to generate long strings of incorrect responses; the probability of being such a student is $0.08$. These students then generate long strings of incorrect responses with a probability that varies by KC. When a student is engaged in this behavior, the probability of a correct response is $0.02$; when a student is not engaged in this behavior, data is generated from the usual two-state BKT model. We refer to this model as BKT with failure sequences {\em (BKT+FS)}.

The data size in each simulation is near the size of the Assistments dataset (section \ref{sec:assist}), with 50 KCs and 3500 students. Each student practiced a random number of KCs, generated by a Poisson distribution with a mean of 5. The number of opportunities for a student to practice each KC also varied randomly, generated by a Poisson distribution with a mean of 8. This means the number of opportunities for practice is statistically independent of the KC. This eliminates the uneven sparsity observed in the real data set. Uniform sparsity makes it possible to use cross-validation over students, and to compare cross-validation to AIC.

For each of the three variations of BKT (classic and two adaptations), we ran 100 simulations. For each of the 300 simulated data sets, we fit 7 models: AFM, PFA with no decay, and R-PFA with 5 different values of the decay weight for the weighted proportion of successes $R_{ij}$: $d = 0.2, 0.4, 0.6, 0.8, 1.0$. For the count of failures, we fixed the decay weight $d=0.1$, since the smallest decay parameter for failures was always optimal for the Assistments data. Complete details for each of the BKT variations are provided in appendix \ref{app:sim}, and code is posted online.\footnote{\url{https://sites.google.com/site/aprilgalyardt/research}} 

To allow model evaluation with cross-validation, we modified the 7 models in two ways from the ones that we fit on Assistments data: we omitted the student effect, and we used fixed rather than random effects for the KC parameters. First, omitting student effects allows us to make predictions for new students. It also implicitly assumes that any new student for whom we will be generating predictions is of average ability. Second, the uneven sparsity in the Assistments data necessitated random effects for KCs and students, but when there is sufficient data at each level, the estimates for random effects from the R function {\tt glmer} and fixed effects from the R function {\tt glm}, used here, will be effectively the same. 

\paragraph{Measures of Model Fit}
On each simulated data set, we compared the 7 models using 3 different measures of model fit: AIC, 5-fold cross-validation (CV) with 0-1 loss (equation \ref{eq:0-1loss}), and 5-fold CV with $L_1$ prediction error loss (PE loss, equation \ref{eq:PEloss}). We omitted the comparison to 0-1 loss with a single test set (i.e., 1-fold CV), because that measure is noisy and unreliable due to high variance of the estimate \cite{James13}.

\subsection{Results and Discussion}
\begin{figure}[htbp]
\begin{center}
\includegraphics[width=\linewidth]{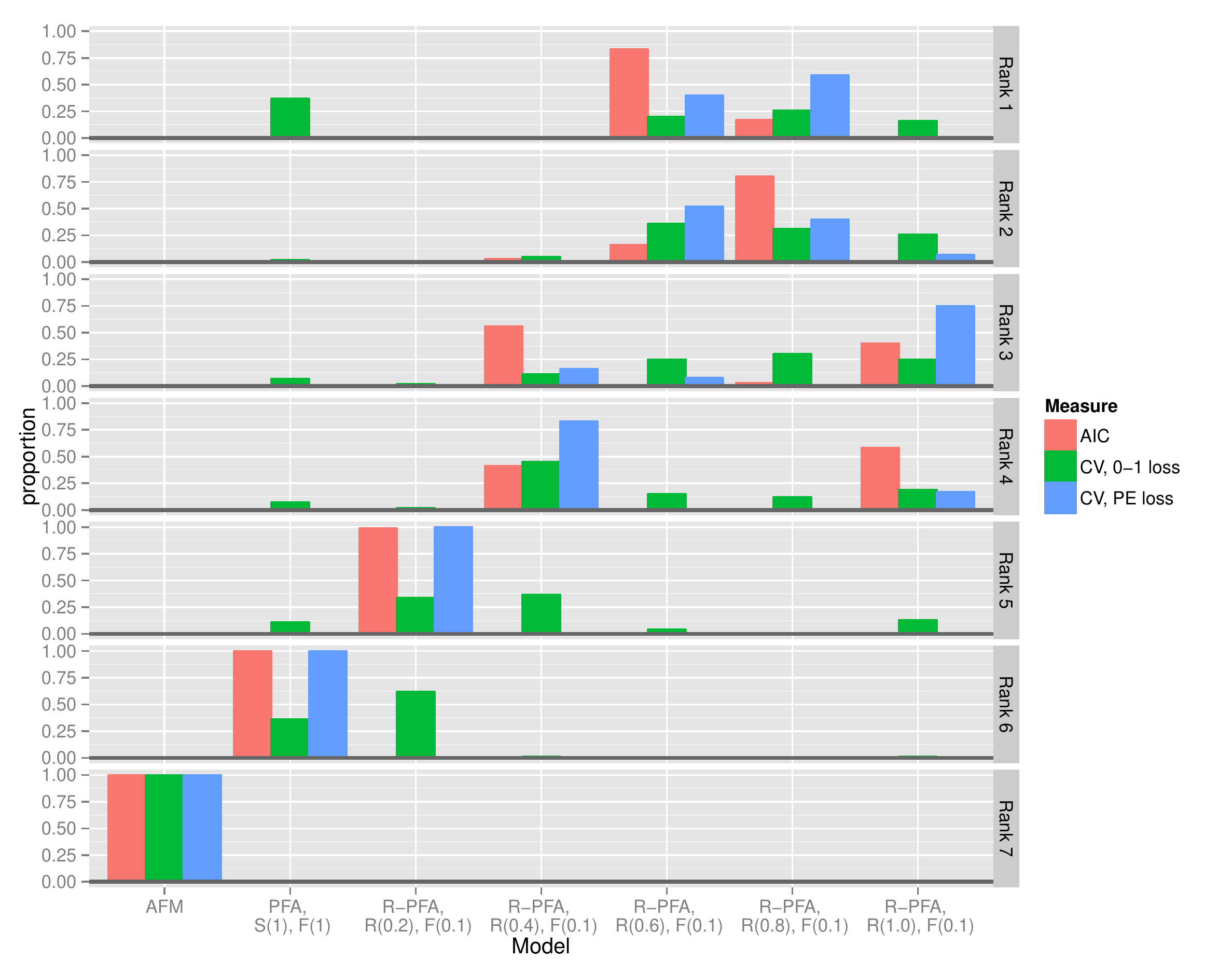}
\caption{Model rankings in the two-state BKT simulation. The rows of graphs indicate the model rankings, e.g., ``Rank 3" indicates that the model was ranked as the third best model in a particular simulation. The height of each bar shows the proportion of simulations where the measure ranked each model. Taller bars at the top of the graph indicate that a model was ranked as a better model. }
\label{fig:sim-bkt-sparse}
\end{center}
\end{figure}

\begin{figure}[htbp]
\begin{center}
\includegraphics[width=\linewidth]{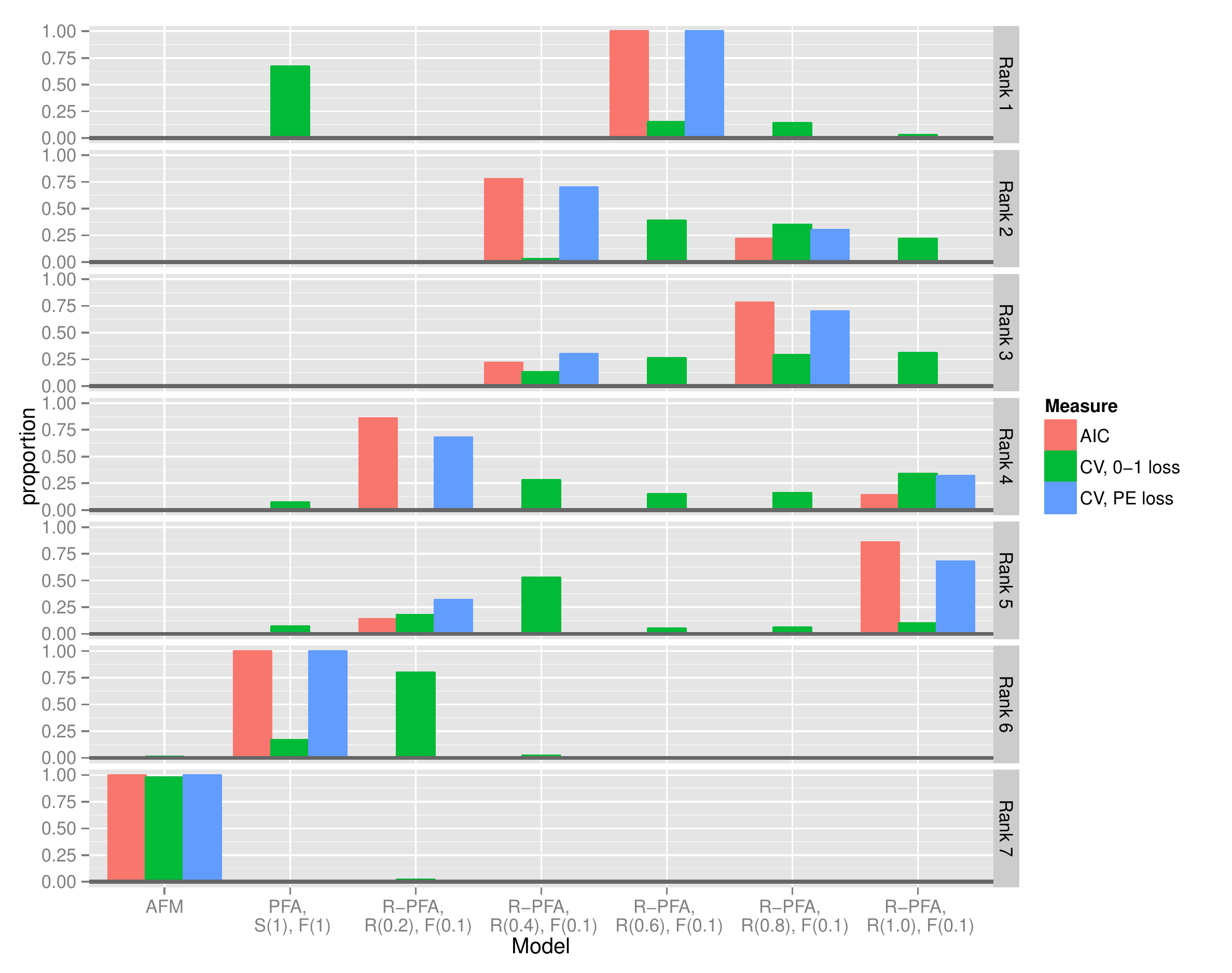}
\caption{Model rankings for each measure in the three-state BKT simulation. The rows of graphs indicate the model rankings, e.g., ``Rank 3" indicates that the model was ranked as the third best model in a particular simulation. The height of each bar shows the proportion of simulations where the measure ranked each model. Taller bars at the top of the graph indicate that a model was ranked as a better model.}
\label{fig:sim-hl-sparse}
\end{center}
\end{figure}

\begin{figure}[htbp]
\begin{center}
\includegraphics[width=\linewidth]{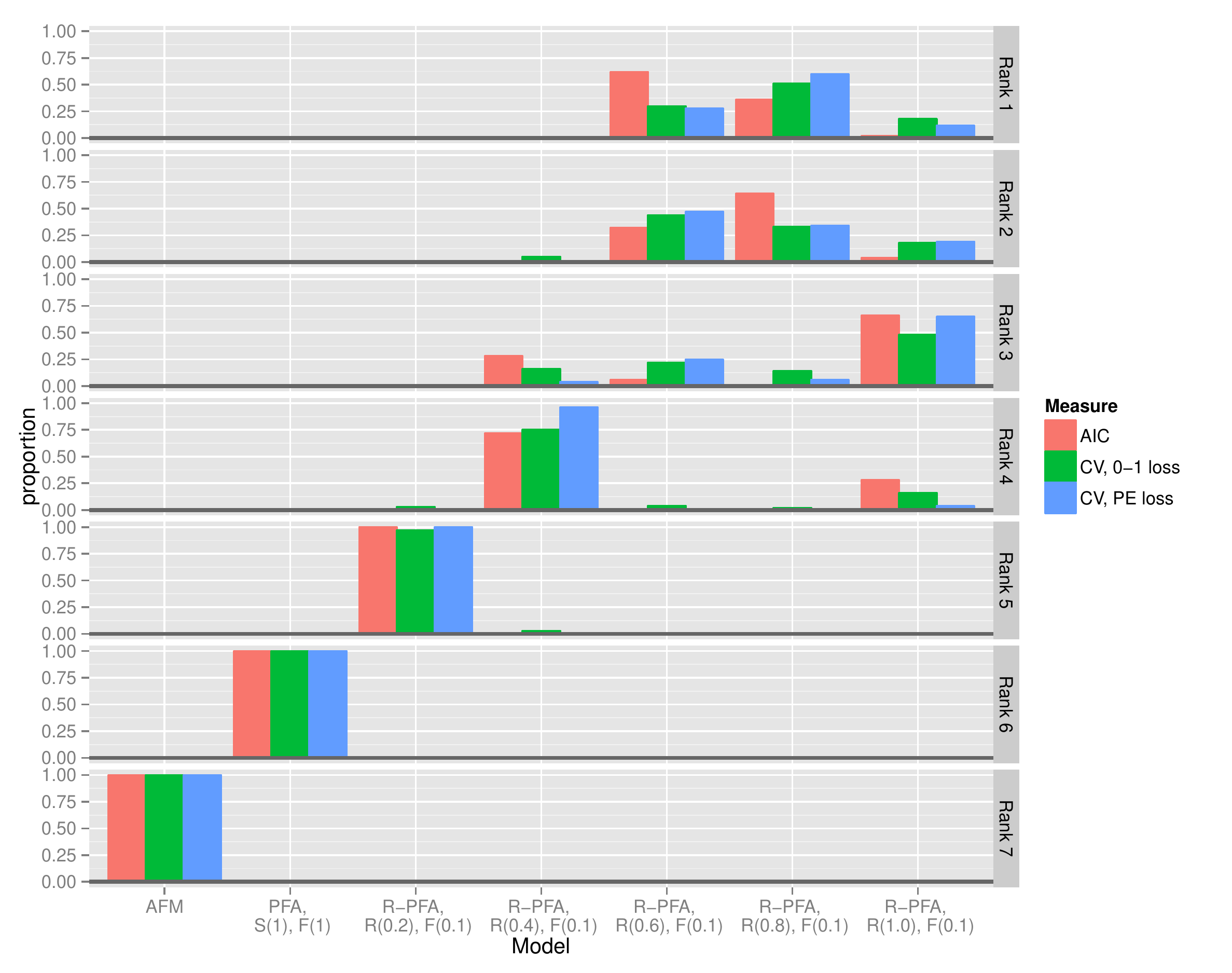}
\caption{Model rankings for each measure in the BKT+FS simulation. The rows of graphs indicate the model rankings, e.g., ``Rank 3" indicates that the model was ranked as the third best model in a particular simulation. The height of each bar shows the proportion of simulations where the measure ranked each model. Taller bars at the top of the graph indicate that a model was ranked as a better model.}
\label{fig:sim-bad-sparse}
\end{center}
\end{figure}

Model rankings for the simulations by each of the goodness-of-fit measures are displayed in figures \ref{fig:sim-bkt-sparse}, \ref{fig:sim-hl-sparse}, and \ref{fig:sim-bad-sparse}. 

\paragraph{Cross-validation with 0-1 Loss has High Variance}
When two models have a very similar fit to the data, measures of model ranking might reasonably diverge in which one they rank as slightly better. For example, in the two-state BTK model in figure \ref{fig:sim-bkt-sparse}: AIC ranks R-PFA($d=0.6$) as the best model in about 80\% of the simulations, and ranks R-PFA($d = 0.8$) in second place. In contrast, CV-PE ranks R-PFA($d$=0.6) as the best model only about 40\% of the time, and puts R-PFA($d$=0.8) in first place in 60\% of the simulations. This is the kind of behavior we expect when two models are similar. Both model measures clearly agree that these are the best two models compared to the rest of the available models. 

By contrast, in about 40\% of the simulations, CV 0-1 ranks PFA as the best model, and in another 40\% of the simulations it ranks PFA in 6th place. This implies that either the PFA regression has highly unstable performance, or that cross-validation with 0-1 loss is an unreliable metric. However, CV with PE loss and AIC are very reliable; they always rank AFM in 7th place, PFA in 6th place, and R-PFA r(0.2) f(0.1) in 5th place.

Moreover, CV with 0-1 loss also fails to reliably rank R-PFA models with the varying decay parameters, while CV with PE loss and AIC do rank them reliably. R-PFA with $d=0.6$ and $d=0.8$ are the highest-ranked (and both very close to the best-performing model on the Assistments data, which had $d=0.7$). R-PFA with bandwidths of 0.4 and 1.0 are not as good. These all outrank R-PFA with $d=0.2$, PFA and AFM in 100\% of the simulations.

The three measures of model fit do not have equal discriminating power. CV with PE loss and AIC rank the models reliablys, and they also largely agree with each other. Cross validation with 0-1 loss is an unreliable measure of model fit.

\paragraph{Model Rankings} 

In all 3 simulation conditions (classic BKT, 3-state BKT, and BKT+FS), in 100\% of the simulations,  the R-PFA models had higher predictive accuracy (judged by AIC and cross-validation with PE loss) than PFA or AFM. For all 3 conditions, the best model was R-PFA with a decay parameter of 0.6 or 0.8. 

In the two-state BKT simulation, AIC and CV-PE rank the models in the same order. CV 0-1 produces ambiguous and unreliable model rankings. R-PFA with decay parameters of 0.6 and 0.8 are the best. R-PFA with bandwidths of 0.4 and 1.0 are not as good. The bandwidth of $d$=0.2 is ranked as the worst R-PFA model in 100\% of the simulations, with PFA in 6th place 100\% of the time, and AFM in last place 100\% of the time. 

In the three-state BKT simulation, once again, CV with 0-1 loss produces ambiguous and unreliable model rankings. According to AIC and CV-PE, R-PFA($d$=0.6) is ranked best 100\% of the time, with $d=0.4$ most often in second place and $d=0.8$ most often in third place. PFA and AFM are again in 6th and 7th place respectively.  

Finally, in the BKT+FS simulation, CV with 0-1 loss agrees with the other measures. R-PFA with $d=0.6$ or $d=0.8$ are ranked as the best models, while PFA is ranked in 6th place in 100\% of the simulations by all three measures. The presence of long strings of incorrect answers, which are produced occasionally by 8\% of the students, makes total number of practice attempts and total number of failures very poor predictors of future success. But because R-PFA only considers recent history, it is not affected by these patterns. A student who was incorrect on the last couple of opportunities for any reason is estimated to have a low probability of responding correctly on the next opportunity. 

The consistent model rankings across all 3 simulation conditions indicates that recent history is a better predictor of learning than a student's full history. The optimal decay parameter range is consistently 0.6-0.8. Thus, the last 3-5 practice opportunities contain sufficient information to judge whether or not a student has learned the KC. 

Simulating from the two- or three-state BKT model offers a best-case scenario for the AFM and PFA models. In a BKT model with no forgetting, the more opportunities that a student has to practice, the more likely it is that a student will transition from the unlearned state to the learned state. Therefore, on average, the total number of opportunities to practice should be proportional to the probability of a correct response. Yet even in this case, R-PFA makes better predictions than PFA. 

When realistic student behavior is added in the BKT+FS simulation, the advantage of R-PFA over PFA becomes even more distinct. The analysis on Assistments data reveals two ways in which the predictions of R-PFA differ from the predictions of PFA: First, R-PFA is better at predicting incorrect answers. Second, the difference between 0-1 loss and prediction error loss indicates that R-PFA has higher confidence in its accurate predictions.

\section{Conclusions}

The primary contributions of this work are:
\begin{itemize}
\item the R-PFA model itself, and its publicly available implementation
\item the comparison of R-PFA to published models and novel baselines on real-world and simulated data, which demonstrates how a student's recent performance history evidences whether or not they have acquired a particular knowledge component
\item the novel visualizations comparing PFA and R-PFA performance, which facilitate logistic regression diagnostics and reveal the source of R-PFA's improvement over alternative models
\item the comparison of measures of model performance, which demonstrates how cross-validation with 0-1 loss is inferior to AIC and to cross-validation with $L_1$ prediction error loss
\end{itemize}

\paragraph{On R-PFA}
R-PFA leverages prior work, including the separation of student and item characteristics (IRT), the grouping of items by skill and the significance of past performance (AFM), the separation of prior successful and unsuccessful practice (PFA), and discounting of older evidence by Gong et al.

R-PFA adds several novel insights to this model evolution. First, a decay-weighted proportion of successes is a better predictor than a decay-weighted count of successes. Second, decay weights should be tuned, rather than determined heuristically (as in the work by Gong et al.). Third, decay weights for successes and failures should be tuned separately. Fourth, it is reasonable and effective to inform the model with a prior "belief" that students who have never attempted the skill will likely fail to answer correctly, e.g., using ghost attempts. 

In aggregate, these insights lead to improvements in predictive accuracy in the true negative rate when recent history contains few correct attempts, and in the true positive rate when recent history mostly consists of correct attempts.

The optimal amount of recent history for modeling is consistent across all of the simulations, and the Assistments data; the best decay parameter for recent successes is consistently $d=\{0.6, 0.7, 0.8\}$. With these decay rates 75-93\% of the weight is on the last 5 attempts, and 55-78\% of the weight is on the last 3 attempts. Thus, empirically, these last 3-5 attempts contain sufficient information about the student's knowledge state to make accurate predictions. Interestingly, this $d$ supports the heuristic, implemented in some adaptive learning systems \cite{heffernan_assistments_2014}, that a student has mastered a skill if a student responds correctly to 3-5 questions in a row on the skill. However, the $R$ predictor is a kind of average that does not require an unbroken streak of correct attempts.

R-PFA is relatively simple mathematically, adding only two tuning weights beyond the parameter structure of PFA. The stability of the decay weights identified in this work implies these weights might be reasonably treated as fixed in new uses of R-PFA, further reducing complexity of R-PFA. R-PFA is more realistic than PFA, because it distinguishes the predictive value of recent performance and older performance. Its predictor and parameter values are easily interpreted, even in the presence of student behaviors that are undesirable or unproductive for learning. Finally, its predictive accuracy improves on PFA. Thus, our ultimate assessment of R-PFA is that it is preferable to PFA and other logistic models in many circumstances where such a model might be used. 

The findings here cast doubt on the validity of the AFM model, because a non-decayed count of successes only, i.e., S-only with $d=1$, outperforms AFM. At present, AFM has uses aside from prediction, including in skill model selection in Learning Factors Analysis \cite{cen_learning_2006}, which may need to adopt different models.

Although we did not compare the predictive accuracy of R-PFA to BKT (although BKT and PFA often have similar accuracy), R-PFA has a strength here as well. Knowledge Tracing is a rather complex Hidden Markov Model. It offers a plausible generative structure for student performance, but {\em notoriously} can fail to return interpretable or accurate estimates of parameters (e.g., \citeNP{beck_identifiability:_2007}). The allure of BKT is the posterior updating of the probability that a student knows the KC after each practice opportunity. R-PFA accomplishes the same thing in simpler way.

Nonetheless, the R-PFA project is by no means complete. In the future, we will consider the relationship of R-PFA to Bayesian Knowledge Tracing, because preliminary work suggests that the two models reveal interesting aspects in each other. We will consider how R-PFA may incorporate richer Q-matrices (multiple skills per item), and how R-PFA may be used to improve cognitive models. We will also extend the R-PFA model with additional predictors, as informed by the comparison to PFA reported above.

\paragraph{On Methods}
This work brings to bear several methodological strengths in terms of model comparison, model fitting, and model analysis.

We evaluated models on both real-world and simulated data. Although simulated data evaluations are rare in the educational data mining literature, they are very popular in statistics. In fact, we argue that real-world datasets have sparse data properties that necessitate both kinds of comparisons.

The model fitting used random effects for all model parameters for students and knowledge components, for both intercepts and slopes. This was necessitated by the sparsity in the Assistments dataset. The random effects were used in both R-PFA and alternative models for a fair comparison.

Our model analysis provides evidence that cross-validation with 0-1 loss, an immensely popular metric in educational data mining, is a poor choice for model comparison. Instead, we argue for the use of AIC as measure of model fit. AIC is seen to be equivalent to cross validation with an $L_1$ prediction error loss. This equivalency is a known result (e.g., \cite{Wasserman04}), but we demonstrate that this holds even when cross validation is making predictions for new students. Any divergence in model choice between AIC and cross validation is due to normal sampling variation, and usually indicates that the fit of the models is similar. A severe divergence in agreement (which we did not see in our simulation) may indicate that the sample size is too small for the model complexity. We note that these conclusions about AIC extend to the Bayesian information criterion (BIC), save that BIC has a higher penalty for model complexity. Cross-validation can be a computationally and time intensive process. AIC and BIC offer a faster and simpler equivalent alternative.

\appendix
\section{Simulation Details}\label{app:sim}

\subsection{Two-state BKT simulation}\label{sec:BKTsim}

This is the usual BKT model, it is a hidden Markov model with an unlearned and a learned state.

\begin{itemize}
\item Knowledge components are indexed $ j = 1, \ldots, K$
\item Students indexed $i = 1, \ldots, N$
\item Student $i$'s response on the $t^{th}$ opportunity to practice KC $j$:
  \[ X_{ijt} =  \left\{\begin{array}{ll} 
      0 & \text{if  incorrect} \\
      1 & \text{if correct} \\
   \end{array}\right.
   \]
\item Denote student $i$'s unobserved knowledge of KC $j$ on the $t^{th}$ opportunity as 
	\[ Z_{ijt} =  \left\{\begin{array}{ll} 
      1 & \text{if  unlearned state} \\
      2 & \text{if learned state} \\
   \end{array}\right.
   \]
\item  Probability of initially knowing KC $j$: $Pr(Z_{ij1} = 2) = \pi_j \sim Beta(1,2)$. 

This distribution has positive probability for all values on the interval [0,1], but is centered at a mean of $\mathbb{E}[\pi_j ] = \frac{1}{3}$. This encapsulates our expectation that in a well-targeted educational intervention, most of the students would not already know the majority of topics which will be taught. The density is shown in figure \ref{fig:beta1-2}.
	
\begin{figure}[h!]
\begin{center}
\includegraphics[width=0.3\linewidth]{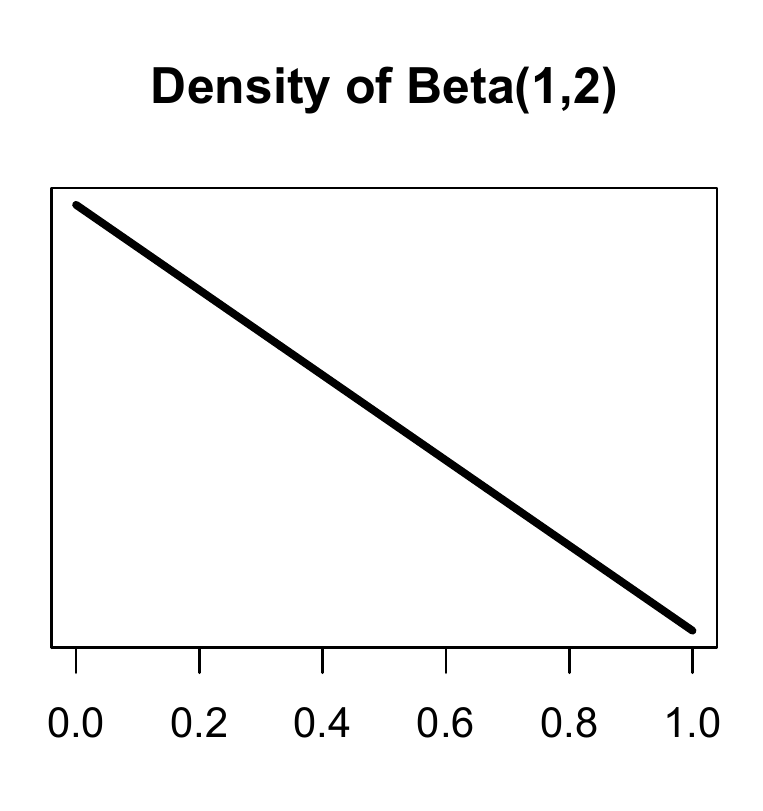}
\caption{Density of $Beta(1,2)$ distribution. }
\label{fig:beta1-2}
\end{center}
\end{figure}

\item Transition matrices for the Markov process are
	\[ P_j =  \left(\begin{array}{cc} 
      1- L_j & L_{j} \\
      0 &1\\
   \end{array}\right)
   \]

$L_j$ is the probability of learning KC k following a practice attempt, generated according to $ L_j \sim Beta(2,2)$. 
 
 This distribution positive probability for all values on the interval [0,1], but is centered at $\mathbb{E}[L_k ] = \frac{1}{2}$.  If $L_k$ is near 1, then a student has a high probability of learning the skill after a single practice attempt. In the same way if $L_k$ is near 0, then a student has a low probability of learning the skill, regardless of how much they practice. This Beta distribution places more probability near 0.5, and lower probability near 0 or 1, reflecting the idea that most students need to practice KCs a couple of times before they learn them. The density is shown in figure \ref{fig:beta2-2}. 
 
 \begin{figure}[h!]
\begin{center}
\includegraphics[width=0.3\linewidth]{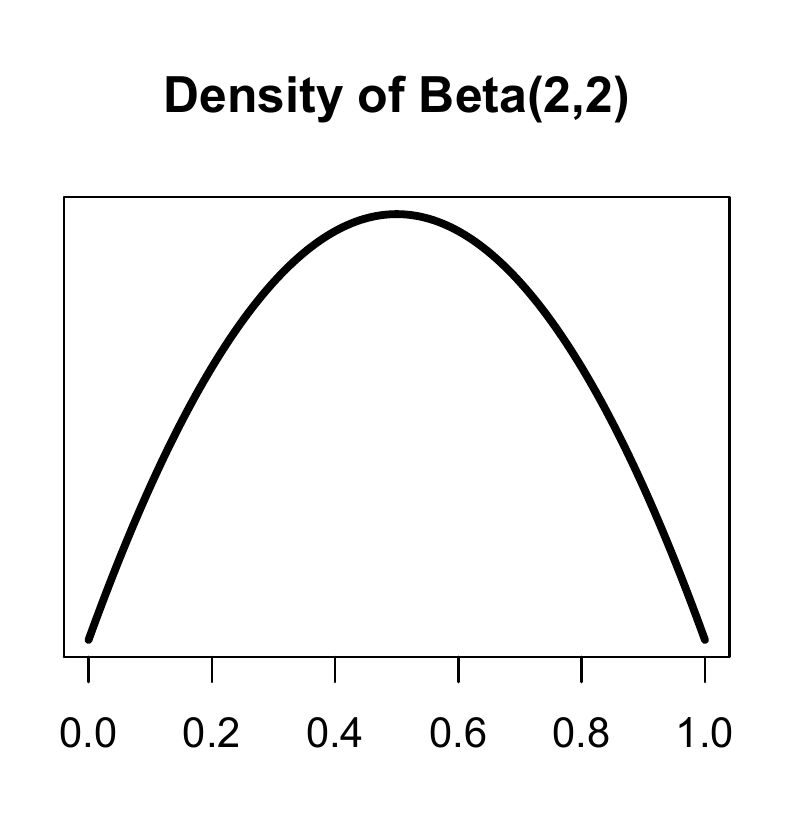}
\caption{Density of $Beta(2,2)$ distribution. }
\label{fig:beta2-2}
\end{center}
\end{figure}
 
 \item Probability of a correct answer in the unlearned state (guessing): $C_{uj} \sim Unif(0.02, 0.3)$
 \[Pr(X_{ijt} = 1 | Z_{ijt} = 1) ~=~ C_{uj}\]
  \item Probability of a correct answer in the learned state (1-slip): $C_{lj} \sim Unif(0.7, 0.98)$
   \[Pr(X_{ijt} = 1 | Z_{ijt} = 2) ~=~ C_{lj}\]
  \item Average number of KC's seen by each student is fixed at {\tt K.n = 5}
  \item Number of KC's seen by student $i$ is generated $J_{i} \sim \min\{K, Poisson(K.n)\}$. 
  \item The KC's that student $i$ answers are drawn without replacement from $\{1, \ldots, K\}$. 
  \item {\tt T.avg = 8} is the average number of practice opportunities for any student on any KC.
  \item The number of practice opportunities for student $i$ on KC $k$ is $O_{ij} \sim \max\{Poisson(T.avg), 2\}$. So that if a student practiced a KC, they practiced it at least twice. 
   
\end{itemize}

\subsubsection{Three-state BKT simulation}\label{sec:3BKTsim}
The 3-state BKT model uses the states unlearned, practicing, and fluent. Students in the unlearned state have a low probability of answering correctly. Students in the practicing state have moderate probabilities of answering correctly. We may think of students in this state as largely understanding the ideas and knowing what to do, but slipping frequently perhaps due to high working memory loads or other causes. Students in the fluent state have a very high probability of answering correctly. 

\begin{itemize}
\item Student $i$'s response on the $t^{th}$ opportunity to practice KC $j$:
  \[ X_{ijt} =  \left\{\begin{array}{ll} 
      0 & \text{if  incorrect} \\
      1 & \text{if correct} \\
   \end{array}\right.
   \]
\item Denote student $i$'s unobserved knowledge of KC $j$ on the $t^{th}$ opportunity as 
	\[ Z_{ijt} =  \left\{\begin{array}{ll} 
      1 & \text{if  unlearned state} \\
      2 & \text{if practicing state} \\
      3 & \text{if fluent state} \\
   \end{array}\right.
   \]

\item Probability for initial states: $\pi_j = (\pi_{j1}, \pi_{j2}, \pi_{j3})$. 
	\begin{eqnarray*}
	P(Z_{ij0} = 1) &= \pi_{j1} &\sim Beta(2,2)\\
	P(Z_{ij0} = 2) & = \pi_{j2} &= 1-\pi_{j1}\\
	P(Z_{ij0} = 3) & = \pi_{j3} &= 0
	\end{eqnarray*}
	This distribution for $\pi_j$ assumes that no student begins practice in the fluent state, so that practice will benefit all students. The $Beta(2,2)$ distribution is shown in figure \ref{fig:beta2-2}. $\pi_{j1}$ can take any value between 0 an 1, but it is more likely to take values nearer to 0.5.  This simulates the idea that for an average KC approximately half the students will start out not knowing the KC at all, and the other half of the students need more practice.

\item Transition matrices for the Markov process are
\[ L_j = \begin{bmatrix}L_{j11} & 1-L_{j11} & 0 \\ 0 & L_{j22} & 1-L_{j22} \\ 0 & 0 & 1 \end{bmatrix}\]
where 
\[L_{j11}, L_{j22} \sim Beta(2,2).\]

With these transition matrices, a student may not transition directly from the unlearned to the fluent state over a single opportunity. However, since this is a 1st order Markov process, it is possible and fairly likely that some students will transition from unlearned to fluent within 2 practice opportunities.

 \item Probability of a correct answer in the unlearned state (guessing): $C_{uj} \sim Unif(0.02, 0.2)$
  \[Pr(X_{ijt} = 1 | Z_{ijt} = 1) ~=~ C_{uj}\]
 \item Probability of a correct answer in the practicing state: $C_{pj} \sim Unif(0.4, 0.7)$
   \[Pr(X_{ijt} = 1 | Z_{ijt} = 2) ~=~ C_{pj}\]
  \item Probability of a correct answer in the fluent state (1-slip): $C_{fj} \sim Unif(0.85, 1)$
     \[Pr(X_{ijt} = 1 | Z_{ijt} = 3) ~=~ C_{fj}\]
 \item Average number of KC's seen by each student is fixed at {\tt K.n = 5}
  \item Number of KC's seen by student $i$ is generated $J_{i} \sim \min(K, Poisson(K.n))$. 
  \item The KC's that student $i$ answers are drawn without replacement from $\{1, \ldots, K\}$. 
  \item {\tt T.avg = 8} is the average number of practice opportunities for any student on any KC.
  \item The number of practice opportunities for student $i$ on KC $k$ is $O_{ij} \sim \max(Poisson(T.avg), 2)$. So that if a student practiced a KC, they practiced it at least twice. 
   
\end{itemize}

\subsubsection{BKT+FS simulation}\label{sec:BKT+fs-sim}
The second adaptation to the familiar 2-state BKT model includes a small proportion of students who occasionally engage in unproductive learning behavior, which produces long strings of incorrect responses, or {\em failure sequences}. This behavior might appear for many different reasons, such as the student engaging in hint-abuse or other gaming behaviors, or the student may simply lack a key prerequisite KC. As a shorthand, we shall refer to students who engage in this behavior as FS-students. 

On each KC, the FS-students will have a probability of engaging in the FS behavior for that KC. The probability that these students will engage in the behaviors depends on the KC, not the student. {\em Whether} a student ever engages in the FS-behavior depends on the student. {\em When} a student does so depends on the KC. 

When a student does engage in FS-behavior, their responses will be a string of primarily incorrect responses with high probability. When a student is not engaging in FS-behavior, data is generated according to an unmodified two-state BKT model.

\begin{itemize}
\item Student $i$'s response on the $t^{th}$ opportunity to practice KC $j$:
  \[ X_{ijt} =  \left\{\begin{array}{ll} 
      0 & \text{if  incorrect} \\
      1 & \text{if correct} \\
   \end{array}\right.
   \]
\item Denote student $i$'s unobserved knowledge of KC $j$ on the $t^{th}$ opportunity as 
	\[ Z_{ijt} =  \left\{\begin{array}{ll} 
      1 & \text{if  unlearned state} \\
      2 & \text{if learned state} \\
         \end{array}\right.
   \]

\item Probability for initial states: $\pi_j = (\pi_{j1}, \pi_{j2})$. Note that $\mathbb{E}[\pi_{j1} ] = \frac{1}{3}$, and the distribution is shown in figure \ref{fig:beta1-2}). 
	\begin{eqnarray*}
	P(Z_{ij0} = 1) &= \pi_{j1} &\sim Beta(1,2)\\
	P(Z_{ij0} = 2) & = \pi_{j2} &= 1-\pi_{j1}
	\end{eqnarray*}

\item Transition matrices for the Markov process are
	\[ P_j =  \left(\begin{array}{cc} 
      1- L_j & L_{j} \\
      0 &1\\
   \end{array}\right)
   \]

$L_j$ is the probability of learning KC $j$ following a practice attempt, generated according to $ L_j \sim Beta(2,2)$. (figure \ref{fig:beta2-2}).

 \item Probability of a correct answer in the unlearned state (guessing): $C_{uj} \sim Unif(0.02, 0.3)$
 \[Pr(X_{ijt} = 1 | Z_{ijt} = 1) ~=~ C_{uj}\]
  \item Probability of a correct answer in the learned state (1-slip): $C_{lj} \sim Unif(0.7, 0.98)$
   \[Pr(X_{ijt} = 1 | Z_{ijt} = 2) ~=~ C_{lj}\]
 
 \item To simulate the FS-behavior:
	\begin{itemize}
	\item For each student draw the indicator $G_i$ for whether student $i$ engages in the FS-behavior,  $G_i \sim Bernoulli(0.08)$. 
	\item For each KC $j$, draw a probability that one of the FS-students will engage in this behavior on this KC.  
	$B_{j}~\sim~ Uniform(0,1).$
	\item Draw an indicator for whether student $i$ will engage in this behavior on KC $j$
	\begin{eqnarray*}
	W_{ij} | G_i = 1 &\sim& Bernoulli(B_j) \\
	W_{ij} | G_i = 0 &=& 0 
	\end{eqnarray*}
	\item If $W_{ij} = 0$, then generate $X_{ij}$ from the 2-state BKT model. 
	\item If $W_{ij} = 1$, then for $t = 1, \ldots, T_{ij}$,  ~~~~~$X_{ijt} | W_{ij} = 1 ~~\sim Bernoulli (0.2)$.

	\end{itemize}
 
 \item Average number of KC's seen by each student is fixed at {\tt K.n = 5}
  \item Number of KC's seen by student $i$ is generated $J_{i} \sim \min(K, Poisson(K.n))$. 
  \item The KC's that student $i$ answers are drawn without replacement from $\{1, \ldots, K\}$. 
  \item {\tt T.avg = 8} is the average number of practice opportunities for any student on any KC.
  \item The number of practice opportunities for student $i$ on KC $k$ is $O_{ij} \sim \max(Poisson(T.avg), 2)$. So that if a student practiced a KC, they practiced it at least twice. 
\end{itemize}

\section*{Acknowledgements}

We thank Neil Heffernan, Ryan Baker, and Yutao Wang for providing the Assistments data set.

\bibliographystyle{acmtrans}
\bibliography{Bib-rPFA}

\begin{thebibliography}{}

\bibitem[\protect\citeauthoryear{Akaike}{Akaike}{1985}]{Akaike85}
{\sc Akaike, H.} 1985.
\newblock Prediction and entropy.
\newblock In {\em A Celebration of Statistics}, {A.~Atkinson} {and}
  {S.~Fienberg}, Eds. Springer: New York, 1--24.

\bibitem[\protect\citeauthoryear{Aleven and Koedinger}{Aleven and
  Koedinger}{2000}]{aleven_limitations_2000}
{\sc Aleven, V.} {\sc and} {\sc Koedinger, K.~R.} 2000.
\newblock Limitations of student control: Do students know when they need help?
\newblock In {\em Intelligent Tutoring Systems}, {G.~Gauthier}, {C.~Frasson},
  {and} {K.~VanLehn}, Eds. Vol. 1839. Springer Berlin Heidelberg, Berlin,
  Heidelberg, 292--303.

\bibitem[\protect\citeauthoryear{Baker, Corbett, Koedinger, and Wagner}{Baker
  et~al\mbox{.}}{2004}]{baker_off-task_2004}
{\sc Baker, R.~S.}, {\sc Corbett, A.~T.}, {\sc Koedinger, K.~R.}, {\sc and}
  {\sc Wagner, A.~Z.} 2004.
\newblock Off-task behavior in the cognitive tutor classroom: when students
  game the system.
\newblock In {\em Proceedings of the {SIGCHI} conference on Human factors in
  computing systems}. {ACM}, 383--390.

\bibitem[\protect\citeauthoryear{Baker, Goldstein, and Heffernan}{Baker
  et~al\mbox{.}}{2011}]{Baker11}
{\sc Baker, R.~S.}, {\sc Goldstein, A.~B.}, {\sc and} {\sc Heffernan, N.~T.}
  2011.
\newblock Detecting learning moment-by-moment.
\newblock In {\em IJAIED}. Vol.~21. 5--25.

\bibitem[\protect\citeauthoryear{Bates, Maechler, Bolker, and Walker}{Bates
  et~al\mbox{.}}{2013}]{lme4}
{\sc Bates, D.}, {\sc Maechler, M.}, {\sc Bolker, B.}, {\sc and} {\sc Walker,
  S.} 2013.
\newblock lme4: Linear mixed-effects models using eigen and s4.
\newblock Computer Program.

\bibitem[\protect\citeauthoryear{Beck and Chang}{Beck and
  Chang}{2007}]{beck_identifiability:_2007}
{\sc Beck, J.~E.} {\sc and} {\sc Chang, K.-m.} 2007.
\newblock Identifiability: A fundamental problem of student modeling.
\newblock In {\em User Modeling 2007}, {C.~Conati}, {K.~McCoy}, {and}
  {G.~Paliouras}, Eds. Number 4511 in Lecture Notes in Computer Science.
  Springer Berlin Heidelberg, 137--146.

\bibitem[\protect\citeauthoryear{Cen, Koedinger, and Junker}{Cen
  et~al\mbox{.}}{2006a}]{Cen06}
{\sc Cen, H.}, {\sc Koedinger, K.}, {\sc and} {\sc Junker, B.} 2006a.
\newblock Learning factors analysis -- a general method for cognitive model
  evaluation and improvement.
\newblock In {\em Proceedings of 8th ITS Conference}. Springer, Berlin /
  Heidelberg, 164--175.

\bibitem[\protect\citeauthoryear{Cen, Koedinger, and Junker}{Cen
  et~al\mbox{.}}{2006b}]{cen_learning_2006}
{\sc Cen, H.}, {\sc Koedinger, K.}, {\sc and} {\sc Junker, B.} 2006b.
\newblock Learning factors analysis -- a general method for cognitive model
  evaluation and improvement.
\newblock {M.~Ikeda}, {K.~D. Ashley}, {and} {T.-W. Chan}, Eds. Vol. 4053.
  Springer Berlin Heidelberg, Berlin, Heidelberg, 164--175.

\bibitem[\protect\citeauthoryear{Cen, Koedinger, and Junker}{Cen
  et~al\mbox{.}}{2008}]{Cen08}
{\sc Cen, H.}, {\sc Koedinger, K.}, {\sc and} {\sc Junker, B.} 2008.
\newblock Comparing two irt models for conjunctive skills.
\newblock In {\em Proceedings of the Proceedings of the 9th International
  Conference on Intelligent Tutoring Systems}, {B.~Woolf}, {E.~Aimer}, {and}
  {R.~Nkambou}, Eds. Springer-Verlag, Montreal, Canada.

\bibitem[\protect\citeauthoryear{Cen, Koedinger, and Junker}{Cen
  et~al\mbox{.}}{2007}]{Cen07}
{\sc Cen, H.}, {\sc Koedinger, K.~R.}, {\sc and} {\sc Junker, B.} 2007.
\newblock Is over practice necessary? --improving learning efficiency with the
  cognitive tutor through educational data mining.
\newblock In {\em Proceedings of the 2007 conference on Artificial Intelligence
  in Education: Building Technology Rich Learning Contexts That Work}. IOS
  Press, Amsterdam, The Netherlands, The Netherlands, 511--518.

\bibitem[\protect\citeauthoryear{Chi, Koedinger, Gordon, Jordan, and
  VanLehn}{Chi et~al\mbox{.}}{2011}]{Chi11}
{\sc Chi, M.}, {\sc Koedinger, K.}, {\sc Gordon, G.}, {\sc Jordan, P.}, {\sc
  and} {\sc VanLehn, K.} 2011.
\newblock Instructional factors analysis: A cognitive model for multiple
  instructional interventions.
\newblock In {\em Proceedings of the 4th EDM Conference}.

\bibitem[\protect\citeauthoryear{Corbett and Anderson}{Corbett and
  Anderson}{1995}]{corbett_knowledge_1995}
{\sc Corbett, A.~T.} {\sc and} {\sc Anderson, J.~R.} 1995.
\newblock Knowledge tracing: Modeling the acquisition of procedural knowledge.
\newblock {\em User Modeling and User-Adapted Interaction\/}~{\em 4,\/}~4,
  253--278.

\bibitem[\protect\citeauthoryear{{de Boeck} and Wilson}{{de Boeck} and
  Wilson}{2004}]{deBoeck04}
{\sc {de Boeck}, P.} {\sc and} {\sc Wilson, M.}, Eds. 2004.
\newblock {\em Explanatory item response models: a generalized linear and
  nonlinear approach.}
\newblock Springer, New York.

\bibitem[\protect\citeauthoryear{Fischer}{Fischer}{1973}]{Fischer73}
{\sc Fischer, G.~H.} 1973.
\newblock The linear logistic test model as an instrument in educational
  research.
\newblock {\em Acta Psychologica\/}~{\em 37}, 359--374.

\bibitem[\protect\citeauthoryear{Galyardt and Goldin}{Galyardt and
  Goldin}{2014}]{galyardt_recent-performance_2014}
{\sc Galyardt, A.} {\sc and} {\sc Goldin, I.~M.} 2014.
\newblock Recent-performance factors analysis.
\newblock In {\em Proceedings of 7th International Conference on Educational
  Data Mining}. London, {UK}.

\bibitem[\protect\citeauthoryear{Gong, Beck, and Heffernan}{Gong
  et~al\mbox{.}}{2011}]{Gong11}
{\sc Gong, Y.}, {\sc Beck, J.~E.}, {\sc and} {\sc Heffernan, N.~T.} 2011.
\newblock How to construct more accurate student models: Comparing and
  optimizing knowledge tracing and performance factor analysis.
\newblock {\em International Journal of Artificial Intelligence in
  Education\/}~{\em 21,\/}~1, 27--46.

\bibitem[\protect\citeauthoryear{Hadfield}{Hadfield}{2010}]{MCMCglmm}
{\sc Hadfield, J.~D.} 2010.
\newblock Mcmc methods for multi-response generalized linear mixed models: The
  {MCMCglmm} {R} package.
\newblock {\em Journal of Statistical Software\/}~{\em 33,\/}~2, 1--22.

\bibitem[\protect\citeauthoryear{Heffernan and Heffernan}{Heffernan and
  Heffernan}{2014}]{heffernan_assistments_2014}
{\sc Heffernan, N.~T.} {\sc and} {\sc Heffernan, C.~L.} 2014.
\newblock The {ASSISTments} ecosystem: Building a platform that brings
  scientists and teachers together for minimally invasive research on human
  learning and teaching.
\newblock {\em International Journal of Artificial Intelligence in
  Education\/}~{\em 24,\/}~4 (Dec.), 470--497.

\bibitem[\protect\citeauthoryear{James, Witten, Hastie, and Tibshirani}{James
  et~al\mbox{.}}{2013}]{James13}
{\sc James, G.}, {\sc Witten, D.}, {\sc Hastie, T.}, {\sc and} {\sc Tibshirani,
  R.} 2013.
\newblock {\em An Introduction to Statistical Learning}.
\newblock Springer Texts in Statistics. Springer.

\bibitem[\protect\citeauthoryear{Kaser, Koedinger, and Gross}{Kaser
  et~al\mbox{.}}{2014}]{kaser_different_2014}
{\sc Kaser, T.}, {\sc Koedinger, K.}, {\sc and} {\sc Gross, M.} 2014.
\newblock Different parameters-same prediction: An analysis of learning curves.
\newblock In {\em Proceedings of 7th International Conference on Educational
  Data Mining}. London, {UK}.

\bibitem[\protect\citeauthoryear{Pavlik, Yudelson, and Koedinger}{Pavlik
  et~al\mbox{.}}{2011}]{Pavlik11}
{\sc Pavlik, P.}, {\sc Yudelson, M.}, {\sc and} {\sc Koedinger, K.} 2011.
\newblock Using contxtual factors analysis to explain transfer of least common
  multiple skills.
\newblock In {\em Artificial intelligence in education}, {G.~Biswas},
  {S.~Bull}, {J.~Kay}, {and} {A.~Mitrovic}, Eds. Vol. 6738. Springer, 256--263.

\bibitem[\protect\citeauthoryear{Pavlik, Cen, and Koedinger}{Pavlik
  et~al\mbox{.}}{2009}]{Pavlik09}
{\sc Pavlik, P.~I.}, {\sc Cen, H.}, {\sc and} {\sc Koedinger, K.} 2009.
\newblock Performance factors analysis --a new alternative to knowledge
  tracing.
\newblock In {\em Proceedings of AIED}. IOS Press, 531--538.

\bibitem[\protect\citeauthoryear{Wasserman}{Wasserman}{2004}]{Wasserman04}
{\sc Wasserman, L.} 2004.
\newblock {\em All of Statistics}.
\newblock Springer Texts in Statistics. Springer.

\bibitem[\protect\citeauthoryear{Wasserman}{Wasserman}{2006}]{Wasserman06nonpar}
{\sc Wasserman, L.} 2006.
\newblock {\em All of Nonparametric Statistics}.
\newblock Springer, New York, NY.

\bibitem[\protect\citeauthoryear{Yudelson, Hosseini, Vihavainen, and
  Brusilovsky}{Yudelson et~al\mbox{.}}{2014}]{yudelson14-javamooc}
{\sc Yudelson, M.}, {\sc Hosseini, R.}, {\sc Vihavainen, A.}, {\sc and} {\sc
  Brusilovsky, P.} 2014.
\newblock Investigating automated student modeling in a java mooc.
\newblock In {\em Proceedings of 7th International Conference on Educational
  Data Mining}. London, {UK}.

\end{thebibliography}

\end{document}